\documentclass[journal]{IEEEtran}
\usepackage{amsmath,amsfonts}
\usepackage{amssymb}

\usepackage{algorithm,balance}
\usepackage{array}
\usepackage[caption=false,font=normalsize,labelfont=sf,textfont=sf]{subfig}
\usepackage{textcomp}
\usepackage{stfloats}
\usepackage{url}
\usepackage{verbatim}
\usepackage{graphicx}
\usepackage{cite}
\usepackage{booktabs}

\PassOptionsToPackage{table}{xcolor}
\PassOptionsToPackage{prologue,dvipsnames}{xcolor}
\usepackage[dvipsnames]{xcolor}
\usepackage[T1]{fontenc}
\usepackage[scaled=1.0]{inconsolata}

\definecolor{cvpink}{RGB}{219, 48, 122}

\usepackage{listings}
\usepackage{pythonhighlight}

\usepackage{tabulary,multirow,overpic}
\usepackage[symbol]{footmisc}
\usepackage{graphicx}

\usepackage{hyperref}
\hypersetup{
    colorlinks=true,
    bookmarksopen=true,
    bookmarksnumbered=true,
    citecolor=blue,
    urlcolor=black,
    pdfborder={0 0 0}
}

\renewcommand{\paragraph}[1]{\noindent\textbf{#1}~}

\definecolor{codegreen}{rgb}{0,0.6,0}
\definecolor{codegray}{rgb}{0.5,0.5,0.5}
\definecolor{codepurple}{rgb}{0.58,0,0.82}
\definecolor{backcolour}{rgb}{0.95,0.95,0.92}

\lstdefinestyle{mystyle}{
    backgroundcolor=\color{backcolour},   
    commentstyle=\color{codegreen},
    keywordstyle=\color{magenta},
    numberstyle=\tiny\color{codegray},
    stringstyle=\color{codepurple},
    basicstyle=\ttfamily\footnotesize,
    breakatwhitespace=false,         
    breaklines=true,                 
    captionpos=b,                    
    keepspaces=true,                 
    numbers=left,                    
    numbersep=5pt,                  
    showspaces=false,                
    showstringspaces=false,
    showtabs=false,                  
    tabsize=2
}

\newcommand{\fref}[1]{Fig.~\ref{#1}}
\newcommand{\sref}[1]{Section~\ref{#1}}
\newcommand{\tref}[1]{Table~\ref{#1}}

\usepackage{threeparttable}

\usepackage{array}
\newcommand{\PreserveBackslash}[1]{\let\temp=\\#1\let\\=\temp}
\newcolumntype{C}[1]{>{\PreserveBackslash\centering}p{#1}}
\newcolumntype{R}[1]{>{\PreserveBackslash\raggedleft}p{#1}}
\newcolumntype{L}[1]{>{\PreserveBackslash\raggedright}p{#1}}

\newcommand{\gTwoO}[0]{g$^2$o}
\usepackage{bm}
\usepackage{tikz}
\usepackage[super]{nth}
\usepackage{algpseudocode}

\begin{document}

\title{Bundle Adjustment in the Eager Mode}
\author{
Zitong Zhan$^1$, 
Huan Xu$^2$, 
Zihang Fang$^3$, 
Xinpeng Wei$^2$, 
Yaoyu Hu$^4$, 
Chen Wang$^{1}$\\ 
\thanks{Corresponding Email: \texttt{\{zitongz, chenw\}@sairlab.org}}
\thanks{$^{1}$Spatial AI \& Robotics (SAIR) Lab, University at Buffalo, NY 14260}
\thanks{$^{2}$Georgia Institute of Technology, GA 30332}
\thanks{$^{3}$Purdue University, IN 47907}
\thanks{$^{4}$Carnegie Mellon University, PA 15213}
}

\markboth{IEEE Transactions on Robotics}%
{Zhan \MakeLowercase{\textit{et al.}}: Bundle Adjustment in the Eager Mode}


\maketitle

\begin{abstract}
Bundle adjustment (BA) is a critical technique in various robotic applications such as simultaneous localization and mapping (SLAM), augmented reality (AR), and photogrammetry. BA optimizes parameters such as camera poses and 3D landmarks to align them with observations. With the growing importance of deep learning in perception systems, there is an increasing need to integrate BA with deep learning frameworks for enhanced reliability and performance. However, widely-used C++-based BA libraries, such as GTSAM, g$^2$o, and Ceres Solver, lack native integration with modern deep learning libraries like PyTorch. This limitation affects their flexibility, ease of debugging, and overall implementation efficiency. To address this gap, we introduce an eager-mode BA library seamlessly integrated with PyTorch with high efficiency. Our approach includes a sparsity-aware auto-differentiation design and GPU-accelerated sparse operations designed for \nth{2}-order optimization.
Our eager-mode BA on GPU demonstrates substantial runtime efficiency, achieving an average speedup of 18.5$\times$, 22$\times$, and 23$\times$ across all benchmarks compared to GTSAM, g$^2$o, and Ceres, respectively. The source code is available at \href{https://github.com/pypose/bae}{\color{cvpink}https://github.com/pypose/bae}.
\end{abstract}

\begin{IEEEkeywords}
Bundle adjustment, Non-linear least squares, Auto-differentiation, Pose graph optimization
\end{IEEEkeywords}

\section{Introduction}

\IEEEPARstart{B}{undle} adjustment (BA) is a fundamental technique in 3D vision, playing a crucial role in various applications such as virtual reality \cite{jiang2024vr-gs}, photogrammetry \cite{he2024dfsfm}, and simultaneous localization and mapping (SLAM) \cite{xu2025airslam}. The primary goal of BA is to refine sensor and environmental parameters, e.g., camera poses and 3D landmarks, so that the parameters are best fitted with observations, e.g., image pixel matching \cite{zhan2024imatching}.

To enhance localization accuracy and preserve semantic information, integrating BA with data-driven methods has become a growing trend \cite{teed2021droid,wang2025imperative, xu2025airslam, zhan2024imatching, fu2024islam, wang2025vggt}. Achieving this often requires implementing BA within deep learning frameworks, such as PyTorch \cite{pt2}, which operates in the \textit{\textbf{eager mode}}\footnote[2]{\textit{Eager mode} in deep learning frameworks such as PyTorch immediately executes the tensor operation when a line of code is called and dynamically constructs a computational graph for automatic differentiation. This mirrors the native Python syntax and provides an intuitive and interactive development experience for researchers, simplifying debugging and prototyping \cite{torch}.}.
Remarkably, the \textit{eager mode} execution has led to the success of PyTorch due to its various advantages such as ease of use and debugging, as well as the flexibility of Python syntax without sacrificing much performance \cite{torch}. 
In contrast, non-eager mode libraries \cite{tensorflow} require users to define a \textit{static} dataflow graph ahead of execution to support differentiation. 
As a result, researchers have shown an overwhelming preference for eager mode programming \cite{he2019mlframeworks}.
Despite these strengths, no existing BA library operates natively in \textit{eager mode} while matching the flexibility and adaptability of deep learning frameworks such as PyTorch, which leads to several drawbacks.

\begin{figure}[t]
    \centering
    \includegraphics[width=1.0\linewidth]{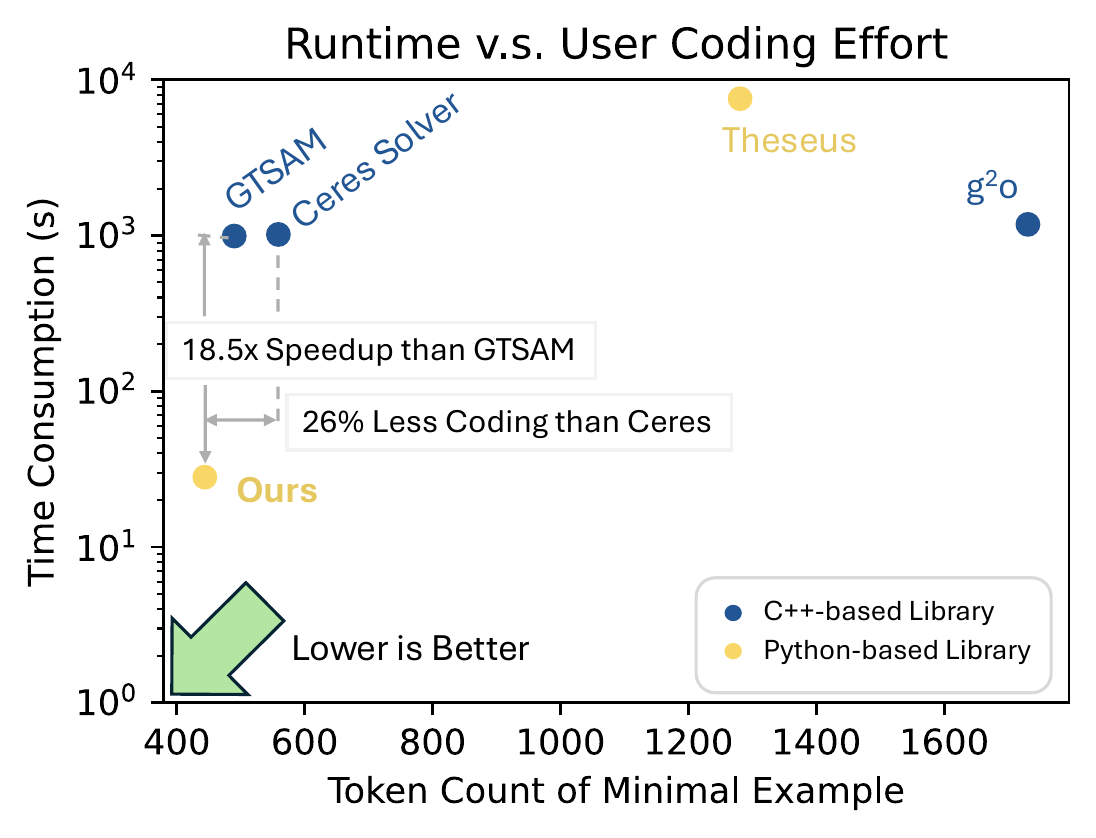}
    \caption{Comparison of runtime and user coding effort across commonly used libraries (GTSAM \cite{gtsam}, Ceres Solver \cite{ceres}, \gTwoO \cite{g2o}, Theseus \cite{pineda2022theseus}). The user coding effort is measured by \textbf{token count} (per GPT-5 tokenization) of each library's \textit{official minimal BA example}, providing a style-agnostic proxy for coding complexity \cite{beizer1990,levitin1986}. For fairness, examples of Ceres, \gTwoO, and ours include the complete user-defined residual computation and compute graph, whereas the GTSAM example uses its built-in library function and omits the explicit camera projection and residual definition. 
    Under this criterion, our method requires \textbf{26\% less code than Ceres}. Time consumption is plotted on a \textit{log scale} and represents a library's total time to optimize BAL \cite{bal} and 1DSfM \cite{wilson_eccv2014_1dsfm} datasets. All libraries are evaluated with consistent CPU and GPU settings and high-performance compiler config detailed in experiments. Both runtime and token count exclude dataset parsing and data preparation.}
    \label{fig:teaser}
\end{figure}

\IEEEpubidadjcol

Existing BA frameworks without eager mode cannot leverage dynamic computational graphs, forcing researchers to rely on static graphs defined ahead of execution. A static graph definition restricts natural and intuitive construction or modification of computational graphs at runtime. This makes dynamic control flows, such as loops, conditionals, and other data-dependent operations, difficult or impossible to implement. Such flexibility is particularly useful in applications like outlier rejection, where runtime decisions depend directly on intermediate optimization results.
Moreover, this limitation significantly impacts the developer experience. Traditional C++-based libraries introduce inefficiencies in development because they do not provide researchers with an intuitive and interactive development environment. In contrast, Python-based libraries inherently support dynamic and flexible behaviors by interpreting user commands on-the-fly. PyTorch \cite{torch} fully embraces this dynamic and Pythonic philosophy by offering simple, consistent, and idiomatic interfaces. This design choice integrates seamlessly with Python’s extensive ecosystem, including debugging, visualization, and data-processing tools. Lastly, traditional BA libraries usually decouple optimization from deep learning frameworks, causing frequent data transfers between CPU-based solvers and GPU-based neural network computations. These transfers incur runtime overhead, especially problematic in large-scale problems. Consequently, enabling BA optimization directly within PyTorch’s eager execution model can substantially reduce these inefficiencies, streamline the research workflow, and provide a more cohesive development experience for researchers.

Nevertheless, building BA frameworks in the \textit{eager mode} is challenging. Core computations, such as \nth{2}-order optimization \cite{robust-bundle-revisited}, sparse Jacobian \cite{bal}, and sparse linear algebra \cite{ceres}, are non-trivial to implement efficiently and scalably within PyTorch's eager execution mode. Unlike traditional C++ libraries, an eager-mode design requires all dataflow to be resolved dynamically at runtime, without relying on pre-defined static factor graphs or symbolic expressions. 
On the other hand, standard PyTorch automatic differentiation (AutoDiff) can only compute dense Jacobians, evaluating gradients between all outputs and parameters even though each BA observation depends only on one camera pose and one 3D landmark. Thus, for \(N\) observations and \(P\) parameters, PyTorch forms a dense \(N\times P\) Jacobian; on the BAL Ladybug scene \cite{bal}, this requires 5.2 TB of memory in double precision, whereas the actual non-zero gradients require only 125 MB, as analyzed in \sref{sec:spjac}.

To tackle these challenges, an optimizer must infer Jacobian sparsity patterns on-the-fly, support dynamic control flow, and remain fully compatible with AutoDiff, all while being designed for scalable GPU execution and seamless integration with GPU deep learning pipelines.
Designing flexible and extensible interfaces to support these operations in eager mode is nontrivial and requires meticulous engineering novelty to balance adaptability, performance, and maintainability.

In this work, we present a new BA library in PyTorch eager mode \cite{pt2}. We introduce sparsity-aware AutoDiff and necessary sparse linear algebra operations in eager mode, addressing \nth{2}-order optimization involving sparse Jacobians of Lie groups. Furthermore, we preserve the original \texttt{optimizer} interfaces of PyTorch, enabling users to leverage the new features with minimal code modifications, consistent with the design philosophy of \cite{wang2023pypose}. As a result, the proposed framework remains highly extensible, allowing users to adapt it easily for other applications such as pose graph optimization (PGO).

To construct sparse Jacobians with minimal user input, we introduce a strategy that automatically traces data manipulation and represents the data flow as a directed acyclic graph.
Additionally, we leverage \texttt{LieTensor} in PyPose to represent the Lie group and Lie algebra for AutoDiff through the batched camera poses.
We adopt the native sparse \texttt{Tensor} in PyTorch to represent the Jacobian, thus avoiding introducing self-defined data structures \cite{gtsam} for ease of use.
We implement new sparse linear solvers and basic math operations in the eager mode, ensuring the entire process of BA is efficient and highly parallelizable. As shown in \fref{fig:teaser}, our method significantly reduces user coding complexity compared to existing libraries while achieving substantial improvements in runtime efficiency.
In summary, our contributions include
\begin{itemize}
    \item We present a new library for BA in the eager mode, showing that optimization traditionally requiring static factor graphs in C++ can easily be carried out in a flexible runtime-built PyTorch computation graph in Python. It works seamlessly with PyTorch autograd engine, allowing learning-based models to be easily combined.
    \item We propose a \textbf{sparsity-aware} AutoDiff framework for \nth{2}-order optimization that leverages automatic computational graph tracing to dynamically identify Jacobian sparsity patterns within directed acyclic graphs, closely aligning with PyTorch's native AutoDiff usage practices. This allows us to represent the resulting sparse Jacobian matrices using native PyTorch sparse tensors, efficiently storing both Euclidean and Lie algebra derivatives. 
    We show that our non-linear optimizers generalize to other problems efficiently, such as PGO.
    \item We implement GPU-accelerated sparse tensor operations in the eager mode. Extensive experiments on both traditional and deep-learning-based structure from motion (SfM) demonstrate the high efficiency of our BA framework on GPU, surpassing GTSAM \cite{gtsam} by 18.5$\times$, \gTwoO\ \cite{g2o} by 22$\times$, and Ceres Solver \cite{ceres} by 23$\times$ in terms of runtime efficiency, even though our eager mode execution trades performance for flexibility. 
\end{itemize}

\begin{figure*}[t]
    \centering
     \includegraphics[width=\textwidth]{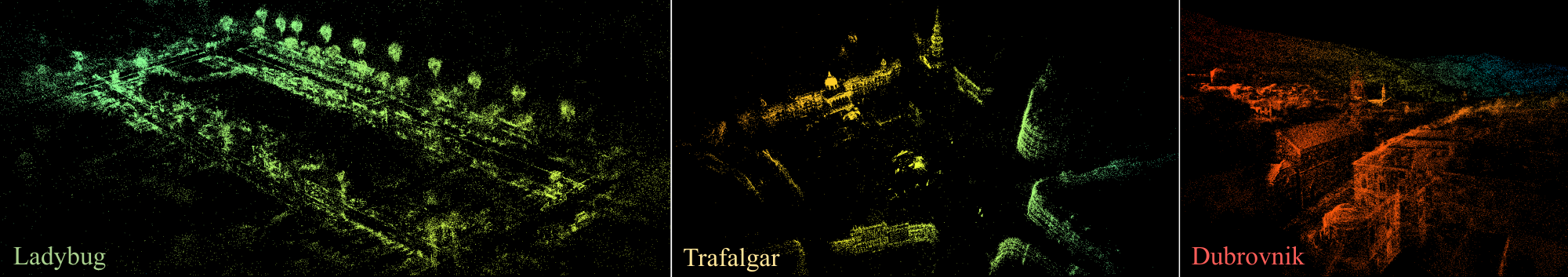}
     \caption{\textbf{Qualitative results on the BAL dataset.} Our method successfully recovered the 3D geometry in the scene. Best viewed digitally.}
     \label{fig:ba_BAL}
\end{figure*}
\section{Related Work}
\subsection{Eager-mode Programming Interfaces}

Eager mode \cite{torch} is an intuitive and user-friendly differentiable programming paradigm in which each tensor operation is executed immediately when the line of code runs, returning concrete values and keeping track of gradient flow, rather than building a computational graph to be executed later. This real-time execution is typically delivered through a Python programming interface, as the programming language has a dynamic nature allowing program behaviors to change at runtime, favored by developers. Eager mode execution by design makes the user code behave like ordinary Python while maintaining differentiability: variables hold actual numbers, dynamic control-flow statements (if, for, while) work naturally, and standard debugging tools or print statements reveal intermediate results at the exact line of code. This immediacy makes experimentation, prototyping, and interactive development straightforward.
On the other hand, non-eager mode \cite{abadi2016tensorflow} involves defining a static computational graph first and then executing the graph as a whole. While this approach enables performance optimization during compilation, it comes at the cost of usability, transparency, and flexibility.

PyTorch \cite{torch} is the first machine learning framework advocating eager mode usage, renowned for its flexible and intuitive programming style.
Its programming style closely mirrors Python, enhancing flexibility and intuitiveness while maintaining simplicity and consistency within the Python ecosystem.
The eager mode design has made PyTorch a favorite among researchers and developers, leading even traditional non-eager mode frameworks to adopt similar programming models \cite{agrawal2019tensorfloweagermultistagepythonembedded}.

\subsection{Factor Graph and Non-linear Optimizers}
Bundle adjustment and SLAM problems are often expressed as factor graphs, in which individual variables (e.g., a camera pose or a 3D landmark) are connected by factors encoding measurement constraints. 
GTSAM \cite{gtsam}, \gTwoO~\cite{g2o}, and Ceres Solver \cite{ceres} all build on static factor‐graph formulations and perform second‐order optimization. All of them offer high-accuracy solutions to BA problems. These libraries are designed for parallel CPU core usage but barely utilize GPUs. 
To achieve end-to-end differentiability under an eager mode interface and for a simpler implementation, PyPose \cite{wang2023pypose} provides a variety of non-linear solvers, including Gauss-Newton and Levenberg-Marquardt (LM), entirely in PyTorch. gradSLAM \cite{jatavallabhula2020slam} is a SLAM demo project that includes an implementation of differentiable Gauss-Newton using PyTorch.
However, they ignored sparsity support \cite{wang2023pypose, jatavallabhula2020slam}, making them impossible to solve even moderate-scale problems. 

To the best of our knowledge, our BA framework provides the first exact eager mode \nth{2}-order optimizer compatible with PyTorch and utilizes sparse data structures for scalability.

\subsection{BA on the GPU}
BA is computationally intensive, especially for large-scale problems involving thousands of images and millions of 3D points; consequently, leveraging GPUs with their massive parallel-processing capabilities has gained traction. Several works have explored GPU-based BA to improve performance while maintaining accuracy.
Ceres Solver \cite{ceres} has introduced limited GPU support for specific operations, such as linear solver, but its primary implementations remain CPU-centric. 
Similarly, DeepLM \cite{huang2021deeplm} attempts to build the LM algorithm partially based on PyTorch. It relies on the autograd engine of PyTorch for calculating the Jacobian values. However, DeepLM is limited to BA problems and does not provide any user interface to define other graph structures such as PGO. Internally, it implements the Jacobian product and the damped linear system using C++ and OpenMP \cite{openmp08} for parallelization on the CPU. Such an implementation choice restricts its compatibility with PyTorch's GPU-accelerated ecosystem and hinders maintainability, making it incompatible with the up-to-date PyTorch 2 \cite{pt2}. 
Consequently, such frameworks fail to fully exploit the capabilities of modern GPU architectures, especially in large-scale BA problems.
In contrast, dedicated GPU-based BA implementations, such as those in PBA \cite{pba}, exploit parallelization for matrix operations and Jacobian computations. However, these solutions are mainly written in customized CUDA kernels designed for older GPUs, limiting their adaptability to new hardware. As a result, PBA is slower than DeepLM, despite that it uses CUDA kernels more extensively. DABA \cite{fan2023daba} is a recent GPU-based BA but focuses on distributed settings, which do not converge to the same error as other methods.

To the best of our knowledge, all the existing GPU-based BA implementations do not support eager mode, therefore hindering their capability in experimentation, debugging, and joint optimization with learning-based models. Moreover, different from previous GPU-based approaches, we introduce LM optimization with sparse tensors, identifying and implementing key sparse linear algebra operations. By encapsulating these operations in standard PyTorch math operators, our approach ensures human interpretability and ease-of-integration and debugging, offering an efficient GPU-based BA solution.

\subsection{BA in Deep Learning}
DROID-SLAM \cite{teed2021droid} is a deep learning-based SLAM system that performs recurrent iterative updates of camera pose and pixel-wise depth through a dense BA layer, enhancing accuracy and robustness. It uses the Gauss-Newton algorithm for simplicity and implements CUDA kernels from scratch specific to its use case. However, it is \textit{not} implemented in eager mode. As a result, its task-specific design is not generalizable to other settings nor easy to extend for other applications.

iMatching \cite{zhan2024imatching} is a self-supervised feature matching learning method that leverages BA as a supervisory signal to enhance the accuracy of feature matching and camera pose estimation. Although based on PyTorch, its BA is implemented with GTSAM \cite{gtsam}, limiting its extensibility and implementation efficiency. In our experiments, we show that our framework can serve as a plug-and-play replacement for traditional BA solvers within such a sophisticated learning framework. 

\begin{figure*}[t]
    \centering
    \includegraphics[width=\linewidth]{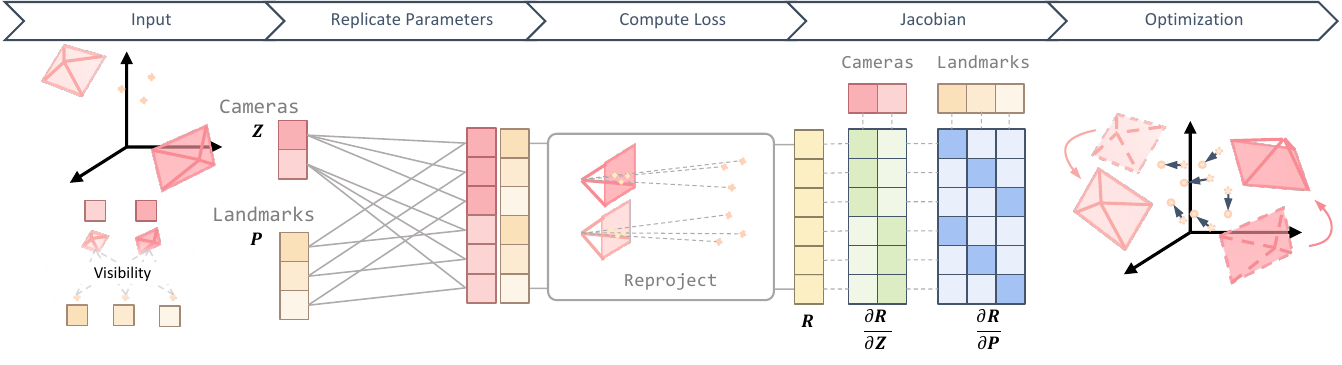}
    \caption{\textbf{Overview of the Bundle Adjustment (BA) process.}  The optimization pipeline starts from the input of camera poses and landmarks, followed by camera reprojection and residual computation. The Jacobian matrix stores the gradient of each residual w.r.t. camera pose and 3D point parameter, showing the contributions of parameters to the Jacobian blocks. The final optimization step iteratively refines camera poses and landmark positions guided by Jacobian.}
    \label{fig:ba_fwd}
\end{figure*}

\section{Preliminaries}
To clarify the challenges of BA in eager mode, we first review non-linear least squares (NLS) optimization, using the Levenberg-Marquardt (LM) algorithm as an example \cite{sh-ch4-diffopt}.

\subsection{Non-linear Least Squares}
BA jointly refines camera parameters and 3D landmarks to minimize reprojection error between the observed 2D image points and the projected 3D points. In practice, BA is often formulated as a non-linear least squares (NLS) problem:
\begin{equation}\label{eq:nls}
\bm{\theta}^* = \arg\min_{\bm{\theta}} \sum_{i=1}^{N} \sum_{j=1}^{M} \lVert \underbrace{\bm{\Pi}(\bm{\zeta}_i, \mathbf{p}_j, \mathbf{K}_i) - \bm{x}_{ij}}_{\mathbf{r}_{ij}} \rVert_2^2,
\end{equation}
where \( \bm{\Pi} \) represents the camera projection model, $N$ is the number of camera poses, $M$ is the number of 3D points, \( \bm{\zeta}_i \in \mathrm{SE}(3) \) is the $i$-th camera pose, \( \mathbf{K}_i \) is camera intrinsic parameters, \( \mathbf{p}_j \in \mathbb{R}^3 \) denotes the $j$-th 3D scene point, and \( \bm{x}_{ij} \) represents the observed 2D pixel location of the 3D point \( \mathbf{p}_j \) in the image of camera \(i\). The goal of the optimization \eqref{eq:nls} is to refine all parameters to minimize the sum of squared reprojection errors $\mathbf{r}_{ij}$, thereby ensuring alignment between the 2D image observations and the 3D geometry.

\noindent \textbf{Notation.} Throughout this paper, we adopt the following notation convention: indexed lowercase symbols (i.e., $\bm{\zeta}_i$, $\mathbf{p}_j$, $\mathbf{r}_{ij}$) denote individual variables, while corresponding aggregated vectors use the uppercase letters in boldface (i.e., $\bm{Z}$, $\bm{P}$, $\bm{R}$). Specifically, camera poses are represented in quaternions, so each $\bm{\zeta}_i \in \mathrm{SE}(3)$. The aggregated camera pose vector is $\bm{Z} = [\bm{\zeta}_1^\top, \ldots, \bm{\zeta}_N^\top]^\top \in \mathbb{R}^{7N}$, the aggregated 3D point vector is $\bm{P} = [\mathbf{p}_1^\top, \ldots, \mathbf{p}_M^\top]^\top \in \mathbb{R}^{3M}$, and the stacked residual vector is $\bm{R} = [\mathbf{r}_{11}^\top, \ldots, \mathbf{r}_{NM}^\top]^\top \in \mathbb{R}^{2NM}$. The total optimization parameters are concatenated as $\bm{\theta} = [\bm{Z}^\top, \bm{P}^\top]^\top \in \mathbb{R}^{7N+3M}$. In practice, visibility is typically sparse due to occlusions or limited fields of view, and the summation in \eqref{eq:nls} is computed only over observed correspondences.

\begin{figure}[t]
\begin{algorithm}[H]
    \caption{The Levenberg-Marquardt algorithm}
    \label{alg:lm}
    \begin{algorithmic}
    
    \Require $\lambda~\text{(damping)}, \bm{\theta}_0~\text{(params)}, \bm{R}~\text{(residuals)}$
    
    \For{$t \gets 1$ to $T$}  
    \vspace{3pt}
    \State $\mathbf{\color{LimeGreen}J} \gets {\dfrac {\partial \bm{R}_{\bm{\theta}_{t-1}}} {\partial \bm{\theta}_{t-1}}}$
    \vspace{3pt}
    \State $\mathbf{\color{LimeGreen}A} \gets \mathbf{\color{LimeGreen}J^\top} \mathbf{\color{LimeGreen}J} $
        \State $\mathbf{\color{LimeGreen}A} \leftarrow \mathbf{\color{LimeGreen}A} + \lambda \cdot \mathrm{diag}(\mathbf{\color{LimeGreen}A})  $
        \State $\Delta \bm{\theta} = \mathrm{solver}(\mathbf{\color{LimeGreen}A}, -\mathbf{\color{LimeGreen}J^\top} \bm{R}_{\bm{\theta}_{t-1}})     $
        \State $\lambda \leftarrow \mathrm{strategy}(\lambda,\text{model information})$
        \State $\bm{\theta}_t \leftarrow \bm{\theta}_{t-1} + \Delta \bm{\theta}             $
    \EndFor\\
    \Return $\bm{\theta}_T$
    \end{algorithmic}
\end{algorithm}
\end{figure}

\subsection{Levenberg-Marquardt Algorithm}

The LM algorithm combines the Gauss-Newton and gradient descent methods to solve an NLS. The LM update rule iteratively adjusts the parameters $\bm{\theta}$ by solving a linear system:
\begin{equation}\label{eq:lm}
\left( \mathbf{J}^\top \mathbf{J} + \lambda \cdot \mathrm{diag} (\mathbf{J}^\top \mathbf{J}) \right) \Delta \bm{\theta} = - \mathbf{J}^\top \bm{R},
\end{equation}
where $\bm{R}$ is the stacked reprojection residual vector,  $\mathbf{J} \doteq \frac{\partial \bm{R}}{\partial \bm{\theta}} \in \mathbb{R}^{(2NM) \times (6N+3M)}$ is the Jacobian of the residuals with respect to the parameters\footnote[3]{The gradient space of $\mathrm{SE}(3)$ has only 6 degrees of freedom.}, \( \lambda \) is a damping factor, and \( \mathrm{diag} (\mathbf{J}^\top \mathbf{J}) \) is a diagonal matrix consisting of the diagonal elements of $\mathbf{J}^\top \mathbf{J}$.
At each iteration, the parameters are updated as 
\begin{equation}
    \bm{\theta}_{t} = \bm{\theta}_{t-1} + \Delta \bm{\theta},
\end{equation}
where the damping factor \( \lambda \) can either be fixed or adjusted based on whether the error is reduced, allowing the algorithm to balance between fast convergence and stability.
A simplified version of the LM method is listed in Algorithm \ref{alg:lm}, whereas a fully implemented version can be found in \cite{pplm}.

\subsection{Sparse Jacobian}
\label{sec:spjac}

Although the LM algorithm \ref{alg:lm} is conceptually simple, the Jacobian matrix $\mathbf{J}$ in a BA problem is often large and has a sparse block structure.
The Jacobian matrix is sparse due to the unique relationship between 3D landmarks and their 2D projections, i.e., the reprojection residual on each 2D pixel $\mathbf{r}_{ij}$ depends only on a single camera pose $\bm{\zeta}_i$ and 3D landmarks $\mathbf{p}_j$, while all other parameters unrelated to the pixel have no gradients \cite{zheng2023distributedbundleadjustmentblockbased}.
As a result, tracking this sparsity pattern is crucial for efficient computation and optimization. {Otherwise, computing every gradient exhaustively is extremely inefficient}\footnote[4]{For example, the sample ``Ladybug'' in the BAL dataset \cite{bal}, with 1,723 camera poses and 156k 3D points, would produce a dense Jacobian consuming 5.2 TB of memory in double-precision floating point and requiring 12 TFLOPs. In contrast, a sparse Jacobian takes only 125 MB in double precision, including block-sparse indexing overhead.}. We next analyze the sparse block pattern.

Since camera poses $\bm{Z} \in \mathbb{R}^{7N}$, 3D points $\bm{P}\in\mathbb{R}^{3M}$, and residuals $\bm{R}\in\mathbb{R}^{2NM}$ are all aggregated vectors comprised of smaller parameter blocks, the Jacobian in BA is in a \textit{sparse block} structure.
A natural way is to partition the large matrix into smaller blocks to highlight the interactions between individual variables, $\bm{\zeta}_i$ and $\mathbf{p}_j$, and their residual $\mathbf{r}_{ij}$ \cite{zheng2023distributedbundleadjustmentblockbased}. In this way, each block is a sub-Jacobian matrix, comprising the partial derivatives of the reprojection error with respect to a specific camera or point parameter. Formally, each residual $\mathbf{r}_{ij}$ is only associated with two Jacobian blocks defined by 
$\mathbf{J}_{\left[\mathbf{r}_{ij}, \bm{\zeta}_i\right]} \doteq \frac{\partial \mathbf{r}_{ij}}{\partial \bm{\zeta}_i} \in \mathbb{R}^{2 \times 6}$ for camera poses
and
$\mathbf{J}_{\left[\mathbf{r}_{ij}, \mathbf{p}_j\right]} \doteq \frac{\partial \mathbf{r}_{ij}}{\partial \mathbf{p}_j} \in \mathbb{R}^{2 \times 3}$ for 3D points.
By only storing such sparse blocks, the space complexity can be reduced from the full-matrix's $\mathcal{O}(n^2)$ to $\mathcal{O}(n)$, where $n=6N+3M$ is the number of parameters to optimize. This space complexity reduction makes it practical for solving large-scale problems.

\section{Methodology}
{Since BA relies on highly sparse Jacobians, which PyTorch’s general AutoDiff system is not designed for, a regular LM implementation \cite{wang2023pypose} and the standard PyTorch AutoDiff pipeline become impractical for BA problems.}
These issues, including constructing the Jacobian with sparsity-aware AutoDiff, sparse linear algebra operations, sparse linear solvers, and their eager mode execution, will be addressed in \sref{sec:jac}, \ref{sec:sparse_ops}, and \ref{sec:linear-solver}, respectively.

\subsection{Constructing a Sparse Jacobian}
\label{sec:jac}
We first introduce the Jacobian matrix storage format and then the construction of the matrix in the eager mode.
We represent the Jacobian tensors using the native PyTorch \texttt{sparse\_bsr} (Block Sparse Row) format \cite{pytorch_sparse_bsr}, which is particularly designed for matrices with a block sparse structure. 
This retains the original form of the Jacobian as a matrix and allows the Jacobian blocks $\mathbf{J}_{\left[\mathbf{r}_{ij}, \bm{\zeta}_i\right]}$ and $\mathbf{J}_{\left[\mathbf{r}_{ij}, \mathbf{p}_j\right]}$ to be directly indexed by their corresponding residual and parameter. For example, the block in the row associated with residual $\mathbf{r}_{ij}$ and column $i$ of the camera Jacobian $\frac{\partial \bm{R}}{\partial \bm{Z}}$ represents the Jacobian block $\mathbf{J}_{\left[\mathbf{r}_{ij}, \bm{\zeta}_i\right]}$, corresponding to residual $\mathbf{r}_{ij}$ and camera $\bm{\zeta}_i$. 
Moreover, compared to other formats such as \texttt{sparse\_coo} (Coordinate List) \cite{wikipedia_sparse_matrix_coo}, it is more efficient for matrix-matrix operations such as $\mathbf{J}^\top \mathbf{J}$ frequently used in LM.
Note that PyTorch lacks support for basic operations for \texttt{sparse\_bsr} format, e.g., the matrix-matrix product. Therefore, we implement all related operations such as matrix-matrix product and matrix diagonal scaling in the eager mode so that the entire LM algorithm can be applied with standard Python operators, which will be discussed in \sref{sec:sparse_ops}.

To better understand the advantages of our method, it is helpful to contrast it with previous approaches.
It is worth noting that previous BA libraries such as GTSAM \cite{gtsam} employ a Jacobian \textit{dictionary}, 
where each block $\mathbf{J}_{\left[\mathbf{r}_{ij}, \bm{\zeta}_i\right]}$ or $\mathbf{J}_{\left[\mathbf{r}_{ij}, \mathbf{p}_j\right]}$ is retrieved by symbolic identifiers, i.e., $(\bm{\zeta}_i, \mathbf{r}_{ij})$ and $(\mathbf{p}_j, \mathbf{r}_{ij})$, the inputs and outputs elements in factor graph.
This design requires explicit symbol tracking to determine the existence of a block. The solver must know which symbols, i.e., $\bm{\zeta}_i$, $\mathbf{p}_j$, participate in each factor $\mathbf{r}_{ij}$ and perform graph searches to determine where each Jacobian block belongs. However, such a representation is fundamentally incompatible with eager-mode frameworks like PyTorch. Eager mode does not maintain symbolic information or a static factor graph; instead, it traces only high-level tensor operations (e.g., arithmetic, copy) on aggregated tensors (e.g., $\bm{Z}$ and $\bm{P}$) during execution. As a result, the autograd engine has no access to the detailed parameter-factor associations required to construct a Jacobian dictionary. Moreover, dictionary storage relies on discrete pointer data structures that are only optimized for CPUs but cannot be parallelized efficiently on GPUs.

{In contrast, for a GPU-friendly Jacobian storage, our method infers block locations directly from runtime and is the first to utilize PyTorch’s native \texttt{sparse\_bsr} format.}
We next introduce the forward pass computation of $\bm{R}$, followed by our sparsity-aware AutoDiff approach in the eager mode, illustrating their combined role in producing the Jacobian $\mathbf{J}$.

\subsubsection{Forward Pass}
\label{sec:forward_pass}
Considering each camera parameter $\bm{\zeta}_i$ and 3D point $\mathbf{p}_j$ influences multiple pixels $\bm{x}_{ij}$ and residuals $\mathbf{r}_{ij}$, we replicate them to match each $\mathbf{r}_{ij}$ with a unique copy of the parameters $(\bm{\zeta}_i$, $\mathbf{p}_j)$ it depends on. 
This simple replication results in a contiguous data layout in memory \cite{pytorch_tensor_indexing} as shown in \fref{fig:ba_fwd}. In the user interface, users only need to index tensors using simple operations like ``\texttt{tensor[indices]}'' on $\bm{Z}$ and $\bm{P}$ along the batch dimension.
The simple tensor usage is further demonstrated in the minimum code example in \sref{sec:min_code}. 

With this structure, each residual now has a dedicated and aligned copy of its corresponding camera and point parameters, enabling batched computation of all camera projections $\bm{x}$. 
This is at the cost of a small one-time memory-copy but in return obtains a contiguous data layout, enabling efficient batched SIMD execution on GPUs.
Moreover, this parallelized and memory-aligned representation of replicated parameters offers a simpler computation of sparse Jacobians in the next stage, which is a dominant factor in BA's computing cost. We next describe how this structure supports our AutoDiff method.

\begin{figure}[t]
    \centering
    \includegraphics[width=1.0\linewidth]{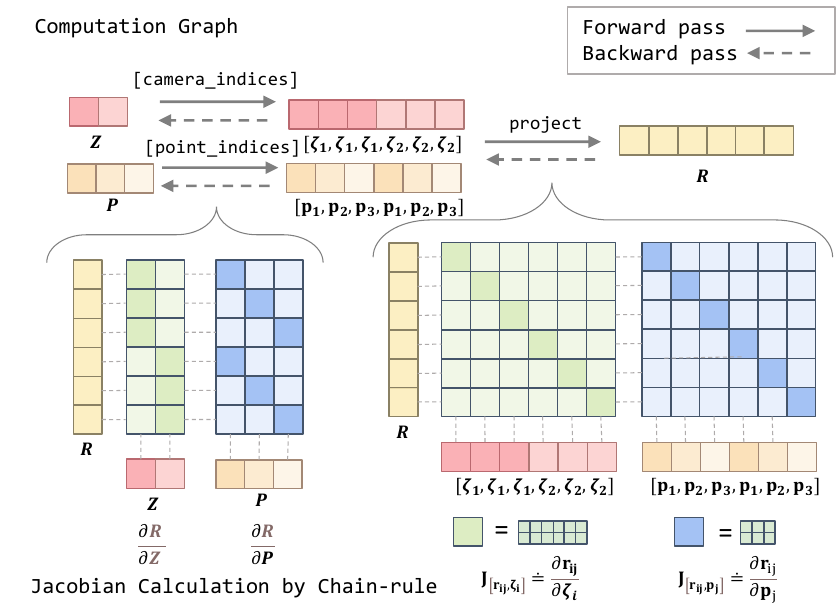}
    \caption{\textbf{Illustration of the sparsity-aware Jacobian construction in BA.} The forward pass is shown by the arrows moving rightward. Each residual \( r_{ij} \) is calculated by parameters of a camera pose $\bm{\zeta}_i$ and a 3D point $\mathbf{p}_j$.  Since each camera and point contribute to multiple reprojections, the parameters are replicated to match the residuals using \texttt{camera\_indices} and \texttt{point\_indices}, respectively. The backward pass, shown by leftward arrows, first propagates through the camera reprojection. Non-zero Jacobian blocks are highlighted with dark color coding. The backpropagation then proceeds through parameter replication. Based on the original cameras and points involved, the blocks are placed in their respective positions.}
    \label{fig:ba_jac}
\end{figure}

\subsubsection{Sparsity-aware AutoDiff in the Eager Mode}
\label{sec:autodiff}
PyTorch's eager mode AutoDiff \cite{pytorch-autograd-tutorial} streamlines gradient computation by eliminating the need for user intervention. In our approach, we aim to achieve the same level of flexibility to efficiently compute sparse Jacobian matrices, which are crucial for many applications. 
The key challenges are two-fold: determining the Jacobian sparsity pattern in $\mathbf{J}$ and computing the non-zero Jacobian blocks $\mathbf{J}_{\left[\mathbf{r}_{ij}, \bm{\zeta}_i\right]}$ and $\mathbf{J}_{\left[\mathbf{r}_{ij}, \mathbf{p}_j\right]}$. The PyTorch autograd engine alone does not have prior information of the sparsity pattern during the forward pass, making it impossible to directly produce $\mathbf{J}$.
To solve this, we propose to use a directed acyclic graph (DAG) to build the computational graph, where nodes represent variables (e.g., tensors, parameters) and edges represent mathematical operations (e.g., addition, multiplication) indicating the data flow.
This graph captures the sequence of operations executed, storing information to support gradient calculations later. 

To construct the sparse Jacobian matrix $\mathbf{J}$, we next analyze the computational graph to distinguish the roles of different operations: tensor indexing operations define the sparsity pattern by establishing dependencies between parameters and output, while arithmetic operations, such as Lie-group multiplication, generate gradients in the non-zero Jacobian blocks.

\paragraph{Determine the sparsity pattern}
The sparsity pattern of the Jacobian matrix $\mathbf{J}$ emerges from the dependencies between individual parameters, $\bm{\zeta}_i$ and $\mathbf{p}_j$, and their corresponding residual $\mathbf{r}_{ij}$. These dependencies, shown in \fref{fig:ba_fwd}, are established by tensor indexing operations performed along the batch dimension during the forward pass. Specifically, the indexing operation assigns each residual $\mathbf{r}_{ij}$ to a unique pair of parameters $\bm{\zeta}_i$ and $\mathbf{p}_j$, determining the precise column locations of the non-zero Jacobian blocks $\mathbf{J}_{[\mathbf{r}_{ij}, \bm{\zeta}_i]}$ and $\mathbf{J}_{[\mathbf{r}_{ij}, \mathbf{p}_j]}$ in the sparse matrix. This assignment ensures that each row of $\mathbf{J}$, which corresponds to a single residual $\mathbf{r}_{ij}$, contains non-zero entries only in the columns associated with the specific camera pose $\bm{\zeta}_i$ and 3D point $\mathbf{p}_j$ that contribute to that residual, thereby defining the sparse structure of the Jacobian. As shown in \fref{fig:ba_fwd}, $\mathbf{r}_{ij}$ has a non-zero block in column $i$ in the camera Jacobian and column $j$ in the landmark Jacobian.

\paragraph{Determine the Jacobian block values}
While the positions of non-zero Jacobian blocks are determined, the actual numerical values of these blocks remain unknown. Our goal is to identify the specific operations generating gradients, which enables our AutoDiff engine to compute only the Jacobian blocks that are necessary. 
Arithmetic calculations (e.g., addition, Lie-group multiplication, camera reprojection denoted as \texttt{project} in \fref{fig:ba_jac}, robust kernel and activation functions) exclusively contribute to computing the numerical values of these Jacobian blocks, in contrast to the tensor indexing operation defining the block placement. These operations transform input parameters into outputs, producing the gradients that populate the Jacobian. Crucially, since they operate independently on each input, they do not affect the overall sparsity structure of the Jacobian matrix. This is because the spatial arrangement of non-zero blocks in the Jacobian is governed by the inputs that are involved in each output. Arithmetic operations do not introduce dependencies between different inputs. As a result, the sparsity structure remains unchanged, even as the values within the non-zero blocks are computed.

Once the sparsity pattern is established, the next step is to compute the values of the non-zero Jacobian blocks efficiently.
Although there's an intuitive solution to traverse through each block and calculate backward gradients one by one, it is inevitably slow. Instead, we compute all blocks within a single backpropagation pass and generate camera and point Jacobian blocks in batch, with stacked shapes of $\mathbb{R}^{(N\times M) \times 2 \times 6}$ and $\mathbb{R}^{(N\times M) \times 2 \times 3}$, respectively. 
The batched Jacobian-value calculation is achieved by composing PyTorch functional programming primitives for batched operations.
Specifically, we use \texttt{func.jacrev} to obtain the Jacobian block calculation for a single pixel residual. We then use \texttt{func.vmap} to apply this computation to the entire batch, generating all Jacobian blocks in parallel. This approach aligns with PyTorch’s SIMD programming style, ensuring both conciseness and efficiency.

In conclusion, there is a clear division of operations' effect on $\mathbf{J}$; indexing operations record block placement, while the rest of the arithmetic operations generate block values. 
This explicit separation ensures efficient calculation of sparse Jacobian structures in our AutoDiff strategy, crucial for computational performance and scalability. 
\textit{It is worth noting that, although this sparsity-aware AutoDiff is inherently complex, users are not required to manage these details.}
We next introduce our approach in automating this process. 

\paragraph{AutoDiff via DAG-Based operation tracking}
Defining large-scale optimization problems with thousands of parameters is typically complex, often requiring manual construction of static compute graphs. Our approach preserves the flexibility of eager mode, handling operation tracking in the backend without requiring users to deal with sparsity details.

At the core of our sparsity-aware AutoDiff system, we propose to use a DAG to dynamically capture computational dependencies during the forward pass.
The DAG is built incrementally and automatically in eager mode execution, with each node representing a tensor and each edge representing an operation that transforms input tensors into output tensors. The graph is initialized with the inputs as the initial nodes. As each operation is executed, it is classified as either tensor indexing or an arithmetic operator and recorded as a directed edge linking the input and output tensors. 

In the backward pass, the DAG is traversed in reverse topological order, allowing gradient signals to propagate from the final output residuals back to their dependent inputs. Each edge in the graph corresponds to a differentiable operation, and for every visited node, its local gradient with respect to each parent is computed and accumulated. Crucially, for arithmetic operations, our aforementioned batched Jacobian values calculation is leveraged to automatically compute the derivative tensors. For tensor indexing operations, our backend extracts the index mappings recorded during the forward pass to determine where the gradients should be routed, effectively preserving the sparsity structure in the reverse direction. \fref{fig:ba_jac} demonstrates the complete process of backpropagation through a BA problem. It first calculates the Jacobian of the camera reprojection, which consists of pure arithmetic operations generating Jacobian block values. When backpropagating the indexing step, the recorded indices are used to place the Jacobian blocks in their correct locations. 

This dynamic DAG construction avoids the need to define an explicit computational graph ahead of time, unlike traditional C++-based optimizers. Users can define their optimization problem using intuitive Python control flows, such as loops and conditional branches, and our system will automatically trace the computation to infer gradient dependencies and the Jacobian sparsity pattern. This makes it possible to construct a dynamic and data-dependent model while retaining full compatibility with GPU-accelerated operations and PyTorch’s eager execution semantics.
As long as the operation is differentiable in PyTorch, its contributions to the Jacobian will be automatically tracked and incorporated by our backend. This design ensures that researchers can prototype rapidly without being constrained by rigid computational graph templates or the need to manually encode sparsity patterns.

In summary, our AutoDiff bridges the gap between classic optimization routines and modern deep learning paradigms. 
This approach not only simplifies the implementation of BA but also generalizes to a variety of optimization tasks such as PGO discussed in \sref{sec:pgo_method}.

\begin{figure}[t]
    \centering
    \includegraphics[width=1.0\linewidth]{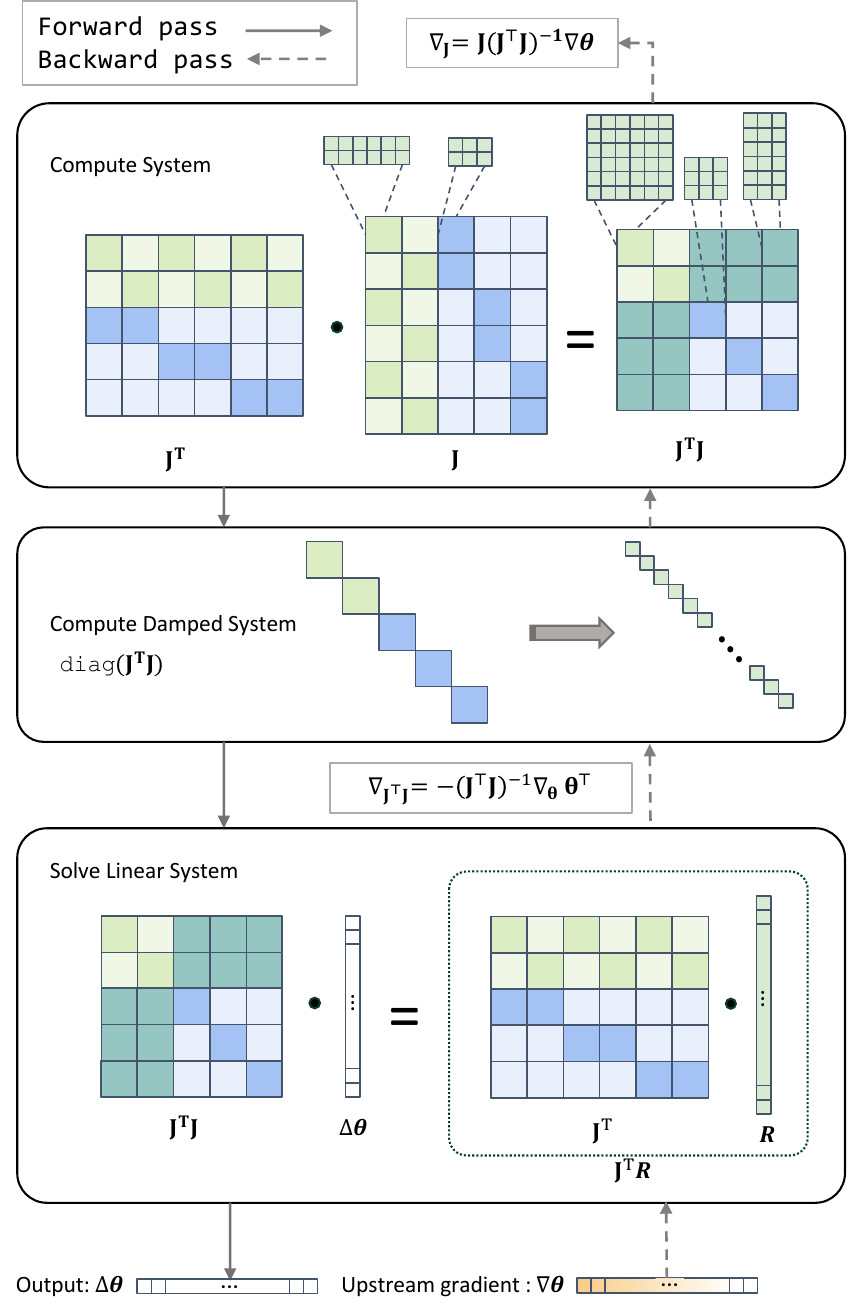}
    \caption{\textbf{Sparse Levenberg-Marquardt Optimization in Eager-Mode Bundle Adjustment.}
The diagram illustrates the core computation steps in the optimization. It starts by forming the normal equations via sparse Jacobian multiplication, followed by computing the damped system using diagonal clamping. The system is then solved via sparse linear solvers, and convergence is checked to determine whether to update parameters or repeat the loop. Our implementation for sparse matrix operations is registered with the native PyTorch operator dispatcher to seamlessly function within the ecosystem.}
    \label{fig:ba_lm}
\end{figure}

\subsection{Basic Sparse Linear Algebra Operations}
\label{sec:sparse_ops}
To complete the remaining steps of the LM algorithm, sparse linear algebra operations are essential. However, PyTorch offers limited support for such operations on sparse tensors. To overcome this limitation, we developed a suite of sparse linear operations. Importantly, unlike libraries such as Ceres Solver, \gTwoO, and GTSAM, which implement operations tailored for their own data structures, our implementation supplies \textit{general-purpose} tensor operations and behaves like native PyTorch mathematical operators, as outlined in \fref{fig:ba_lm}. For example, users can apply sparse matrix-matrix multiplication directly using the familiar ``\texttt{@}'' syntax without rewriting their code, while still benefiting from the memory and speed advantages of sparsity. This design ensures our operators remain both human-interpretable and integrable into a wide range of PyTorch applications with ease.

\subsubsection{Matrix Multiplication}

Matrix multiplication plays a critical role in computing the Jacobian multiplication $\mathbf{J}^\top \mathbf{J}$, where $\mathbf{J}^\top$ is a sparse matrix in Compressed Sparse Row (CSR) \cite{pytorch_sparse_csr} or Block Sparse Row (BSR) format. The multiplication of two sparse matrices, commonly known as Sparse General Matrix-Matrix Multiplication (SpGEMM) \cite{two-fast-algo}, is essential for this computation. SpGEMM is inherently more complex than its dense counterpart due to irregular memory access and variable sparsity patterns. We separate this operation into two phases: symbolic searching and numerical multiplication.

Symbolic searching identifies the sparsity structure of the output matrix without computing any actual values \cite{dalton2015optimizing}. Intuitively, this step determines which entries in the $\mathbf{J}^\top \mathbf{J}$ will be non-zero, based solely on the positions of non-zero elements in $\mathbf{J}^\top$ and $\mathbf{J}$. Essentially, it constructs a multiplication table, a blueprint indicating which blocks from the input matrix should be multiplied together and where in the output matrix each result should be stored. For example, if a non-zero block in row $i$, column $k$ of $\mathbf{J}^\top$ and another in row $k$, column $j$ of $\mathbf{J}$ exist, symbolic search determines that the product of these two blocks contributes to the entry at position $(i, j)$ in the result $\mathbf{J}^\top \mathbf{J}$. 
The precomputed blueprint allows the algorithm to know exactly which computations are needed. 
With this, the numerical multiplication phase can batch all necessary computations to exploit the high parallelism of the GPU. 

This symbolic structure depends only on the sparsity pattern of the input matrix, regardless of the numerical values inside. As such, it remains constant across all iterations of LM because the sparsity pattern of $\mathbf{J}$ doesn't change. We compute the symbolic search only once in the first iteration. The multiplication table is then cached and reused throughout the optimization, avoiding expensive recomputation and greatly improving runtime efficiency in iterative solvers like LM.

We implement the CSR multiplication using the cuSPARSE library \cite{cusparse}, leveraging its optimized routines for sparse linear algebra operations. For the BSR format, we develop custom CUDA kernels and utilize Warp \cite{warp2022} to handle the block-structured sparsity efficiently.
These sparse matrix operations are registered to the PyTorch operator dispatcher. Users can perform matrix multiplication using the PyTorch syntax ``\texttt{mat1 @ mat2}'' with previously unsupported sparse types. This design choice ensures that no modifications to the native eager mode API or user code are necessary, preserving the ease of use while adding new sparse linear algebra capabilities.

\subsubsection{Matrix-Vector Product}

The matrix-vector product is for computing $\mathbf{J}^\top \bm{R}$, where $\mathbf{J}^\top$ is sparse while $\bm{R}$ is a dense vector. This operation, commonly referred to as Sparse Matrix-Vector product (SpMV) \cite{matrix-computations}, is one of the few sparse operations natively supported by PyTorch, which internally utilizes the cuSPARSE library for efficient computation.

\subsubsection{Diagonal Clamping and Scaling}

Diagonal clamping and scaling are essential operations in various numerical algorithms, especially when adjusting the diagonal elements of a matrix for stability or regularization purposes. These operations are represented by functions such as $\mathrm{diagonal\_clamp}(\mathrm{min}, \mathrm{max})$, which clamps the diagonal elements within a specified range to ensure numerical stability, and $\lambda \cdot \mathrm{diag}(\mathbf{A})$, which scales the diagonal elements of matrix $\mathbf{A}$ by a factor $\lambda$.
Since PyTorch lacks native support for these operations with sparse matrices, we implemented them using custom Triton kernels \cite{triton_github}. This allows diagonal clamping and scaling directly on sparse matrices. 

\subsection{Sparse Linear Solvers} \label{sec:linear-solver}

Sparse linear solvers play a crucial role in computing parameter updates within the LM algorithm. The task involves solving a linear system of the form $\mathbf{A}\bm{x} = \bm{b}$, which is commonly expressed in code as \texttt{x = solver(A, b)}. In the context of LM, the coefficient matrix $\mathbf{A} = \mathbf{J}^\top \mathbf{J} + \lambda \cdot \mathrm{diag}(\mathbf{J}^\top \mathbf{J})$ is a sparse symmetric positive-definite (SPD) matrix that typically contains millions of non-zeros but less than 0.01\% density, and the right-hand side vector $\bm{b} = -\mathbf{J}^\top \bm{R}$ is dense. The goal is to find the update $\Delta \bm{\theta}$ by solving the linear system $\mathbf{A} \Delta\bm{\theta} = \bm{b}$, as required by~\eqref{eq:lm}.
Linear solvers are generally classified into direct and iterative methods. Each has distinctive characteristics in terms of complexity and scalability, and we provide the following two solver implementations to address varying problem scales: Sparse Direct Solver and Sparse Preconditioned Conjugate Gradient (PCG) Solver to handle small- and large-scale systems, respectively. 
\subsubsection{Sparse Direct Solver (Cholesky Factorization)}

Direct solvers provide exact solutions by decomposing the matrix $\mathbf{A}$ into simpler factorized forms. Among various direct methods, Cholesky factorization \cite{cholesky} is preferred for semi-positive definite (SPD) matrices. Cholesky decomposition factors the matrix $\mathbf{A}$ as
$
\mathbf{A} = \mathbf{L}\mathbf{L}^\top
$,
where $\mathbf{L}$ is a lower triangular matrix. This decomposition allows solving the linear system through a finite number of forward and backward substitution steps. It is highly efficient for small to medium-sized problems.

A key feature of sparse direct solvers is symbolic factorization, a preprocessing stage that identifies the sparsity pattern of the factorization without numeric values. It involves constructing an elimination tree and analyzing the dependencies among variables. Although internally complicated, it is independent of numerical values and only needs to be computed once for matrices with static sparsity patterns.

To optimize performance, we implement a caching strategy similar to the symbolic search in sparse matrix multiplication. Specifically, we perform symbolic factorization during the first LM iteration and cache its results. Subsequent LM iterations reuse the cached symbolic factorization pattern, performing only the numerical factorization, thereby saving computational costs.
This ensures minimal runtime overhead from redundant symbolic factorization steps. Our implementation leverages GPU-accelerated sparse Cholesky factorization routines \cite{nvidia_cudss_2025}, fully exploiting parallelism for numerical factorization.

\subsubsection{Iterative Sparse Solver (Preconditioned Conjugate Gradient)}

While direct solvers are efficient for smaller systems, iterative methods scale better with large-scale systems due to their lower memory footprint and computational complexity per iteration. Among iterative methods, the Preconditioned Conjugate Gradient (PCG) algorithm is particularly well-suited for large-scale SPD systems \cite{block_pcg}.

The PCG method iteratively refines solutions by repeatedly applying SpMV and scalar operations. At each iteration, PCG computes the residual, updates search directions, and refines the solution vector $\Delta\bm{\theta}$ until convergence criteria are met. However, PCG convergence heavily depends on matrix conditioning. To enhance convergence, we adopt a diagonal preconditioning strategy in \cite{bal}. The preconditioner, a diagonal matrix, approximates $\mathbf{A}^{-1}$, improving the condition of the linear system and reducing iterations. 
Additionally, we implement the PCG using CUDA graph capture for further acceleration, with details presented in Appendix \ref{x:cudagraph}.

\subsubsection{Unified and Extensible API}
In both types of solvers, our implementation focuses on concise sparse operators and maintains compatibility with the existing PyPose API originally for dense linear solvers \cite{ppsolver}. This design choice ensures that our optimizer is easy to deploy in research settings and can be readily extended to accommodate more complex non-linear optimization strategies. Users remain agnostic to these sophisticated backend optimizations. By fully exploiting GPU throughput, our approach enables the LM algorithm to be seamlessly executed in the eager execution mode, allowing for straightforward and efficient code development.

\subsection{Generalization to Other Optimization Problems}
\label{sec:pgo_method}

To demonstrate the versatility of our framework, we show that it naturally extends beyond BA to another sparse non-linear least squares problem, PGO. PGO aims to estimate a set of camera poses \(\bm{Z} = [\bm{\zeta}_1^\top, \ldots, \bm{\zeta}_N^\top]^\top\) where each \(\bm{\zeta}_i \in \mathrm{SE}(3)\), given relative pose measurements between pairs of cameras, often derived from odometry or loop closure constraints. Unlike BA, which jointly optimizes camera poses and 3D landmarks, PGO focuses solely on pose refinement but is still a computationally demanding task.
Formally, PGO can be formulated as a weighted non-linear least squares,
\begin{equation}\label{eq:pgo}
\bm{Z}^* = \arg\min_{\bm{Z}} \sum_{(i, j) \in \mathcal{E}} \lVert \bm{\Omega}_{ij}^{1/2} \mathbf{r}_{ij}(\bm{\zeta}_i, \bm{\zeta}_j, \mathbf{T}_{ij}) \rVert_2^2,
\end{equation}
where \(\mathcal{E}\) is the set of edges representing pairwise constraints, \(\mathbf{T}_{ij} \in \mathrm{SE}(3)\) is the measured relative pose between cameras \(i\) and \(j\), \(\bm{\Omega}_{ij} \in \mathbb{R}^{6 \times 6}\) is the information matrix encoding measurement uncertainty, and \(\mathbf{r}_{ij}\) is the residual defined on the Lie algebra \(\mathfrak{se}(3)\),
\begin{equation}
\mathbf{r}_{ij} = \log(\bm{\zeta}_i^{-1} \bm{\zeta}_j \mathbf{T}_{ij}^{-1}),
\end{equation}
where \(\log(\cdot)\) maps the relative pose error from the Lie group \(\mathrm{SE}(3)\) to its tangent space \(\mathbb{R}^6\).

To minimize the above residual, our framework adapts seamlessly to PGO by leveraging the same core components developed for BA. Similar to BA's forward pass described in \sref{sec:forward_pass}, PGO requires replicating node parameters to match each edge constraint: given an edge \((i, j) \in \mathcal{E}\), the corresponding poses \(\bm{\zeta}_i\) and \(\bm{\zeta}_j\) are retrieved via tensor indexing on $\bm{Z}$ along the batch dimension. The indexing operations automatically determine the sparsity pattern of the Jacobian matrix \(\mathbf{J} = \frac{\partial \bm{R}}{\partial \bm{Z}}\), where each residual \(\mathbf{r}_{ij}\) produces non-zero blocks only in columns corresponding to poses \(\bm{\zeta}_i\) and \(\bm{\zeta}_j\).

The residual computation involves Lie group operations, including inversion, composition, and logarithmic mapping. All of the operations are differentiable arithmetic operations in our DAG-based AutoDiff system (\sref{sec:autodiff}). These operations generate the numerical values of the Jacobian blocks through batched backpropagation, while preserving the sparsity structure established by the indexing step. All derivatives are computed on the Lie manifold using \texttt{LieTensor} in \cite{wang2023pypose}, ensuring geometrically consistent gradients without manual derivation. The resulting sparse Jacobian in \texttt{sparse\_bsr} format is then processed by the same sparse linear algebra operations (\sref{sec:sparse_ops}) and linear solvers (\sref{sec:linear-solver}) used in BA, with the LM algorithm applied without modification. This demonstrates that our framework provides a unified infrastructure for sparse optimization problems beyond BA.

\section{Minimum Runnable Code}
\label{sec:min_code}
To illustrate the practical usage of our framework, we provide a concrete implementation of BA in the eager mode. We emphasize that, despite the sophisticated backend mechanisms introduced in the previous sections, the user-facing API remains simple and intuitive. Due to the minimal changes in API usage, users can re-use the same code style of dense LM provided by PyPose \cite{wang2023pypose} for our new sparse LM. A minimum runnable code example for BA in the eager mode with 1 camera and 8 points is listed below. The \texttt{pp.Parameter} wrapper (line 12-13) enables our sparsity-aware AutoDiff by automatically tracing tensor indexing operations to determine the Jacobian sparsity pattern, as described in \sref{sec:autodiff}. The \texttt{psjac} decorator marks the batched residual function as independently applied to each batch sample, accelerating sparse Jacobian tracing and assembly without changing the function behavior. The index tensors \texttt{cidx} and \texttt{pidx} (line 16-17) specify camera-to-observation and point-to-observation correspondences, respectively, which are used to replicate parameters for batched residual computation as described in \sref{sec:forward_pass}. Specifically, \texttt{cidx[k]} indicates the camera ID that observes the $k$-th observation, and \texttt{pidx[k]} indicates the 3D point ID corresponding to the $k$-th observation. To automatically balance the convergence rate and stability, a trust region strategy \texttt{TrustRegion} (line 32) in \cite{wang2023pypose} can be applied to dynamically adjust the damping factor $\lambda$ based on the ratio between actual and predicted loss reduction.

\begin{python}[]
import torch
import pypose as pp
from torch import nn
from pypose.optim import LM
from pypose.optim.solver import PCG
from pypose.optim.strategy import TrustRegion
from pypose.optim.scheduler import StopOnPlateau
from pypose.autograd.function import psjac

class ReprojErr(nn.Module):
    def __init__(self, poses, points):
        super().__init__()
        # sjac: enabling tracing of sparse Jacobian
        self.poses = pp.Parameter(poses, sjac=True)
        self.points = pp.Parameter(points, sjac=True)

    @psjac  # parallelize assembly of sparse Jacobian
    def project(poses, points):
        points = poses.Act(points)
        return - points[..., :2] / points[..., [2]]

    def forward(self, pixels, cidx, pidx):
        poses = self.poses[cidx]
        points = self.points[pidx]
        return ReprojErr.project(poses, points) - pixels

torch.set_default_device("cuda")
npts, poses = 8, pp.randn_SE3(1)
points = torch.randn(npts, 3)
points[:, 2] += 4  # positive depth
cidx = torch.zeros(npts, dtype=torch.long)
pidx = torch.arange(npts)
pixels = torch.randn(npts, 2)
inputs = (pixels, cidx, pidx)

model = ReprojErr(poses, points)
solver = PCG(tol=1e-4, maxiter=250)
strategy = TrustRegion(up=2.0, down=0.5**4)
optimizer = LM(model, solver, strategy, sparse=True)
scheduler = StopOnPlateau(optimizer, steps=5, verbose=True)

while scheduler.continual():
    loss = optimizer.step(inputs)
    scheduler.step(loss)
\end{python}

\begin{table*}[t]
    \centering
    \setlength{\tabcolsep}{0.26em}
    \vspace{5.5pt}
    \renewcommand{\arraystretch}{1.2}
    \caption{Performance comparison with CPU-based BA frameworks on the BAL dataset.}
    \begin{tabular}{L{0.1\linewidth}|C{0.06\linewidth} C{0.06\linewidth} C{0.06\linewidth}|C{0.055\linewidth} C{0.055\linewidth}|C{0.055\linewidth} C{0.055\linewidth}|C{0.055\linewidth} C{0.055\linewidth}|C{0.055\linewidth} C{0.055\linewidth}|C{0.055\linewidth} C{0.055\linewidth}}
    \toprule
    \multirow{2}{*}{Scene}	&	\multirow{2}{*}{Camera}	&	\multirow{2}{*}{Points}	&	\multirow{2}{*}{Pixels}	&	\multicolumn{2}{c|}{GTSAM \cite{gtsam}}	&	\multicolumn{2}{c|}{\gTwoO~\cite{g2o}}	&	\multicolumn{2}{c|}{Ceres \cite{ceres}}	&	\multicolumn{2}{c|}{\textbf{Ours} (PCG)}	&	\multicolumn{2}{c}{\textbf{Ours} (Cholesky)}		\\
     & & & & Time$\downarrow$	&	Error$\downarrow$	&	Time$\downarrow$	&	Error$\downarrow$	&	Time$\downarrow$	&	Error$\downarrow$	&	Time$\downarrow$	&	Error$\downarrow$	&	Time$\downarrow$	&	Error$\downarrow$	\\
     \midrule
    Ladybug	&	1723	&	156502	&	678718	&	12.43	&	2.540	&	59.12	&	1.313	&	177.15	&	1.146	&	\textbf{1.60}	&	\textbf{1.120}	&	6.01	&	1.134	\\
    Trafalgar	&	257	&	65132	&	225811	&	8.47	&	0.896	&	7.25	&	0.863	&	13.41	&	0.856	&	5.81	&	0.854	&	\textbf{1.27}	&	\textbf{0.853}	\\
    Dubrovnik	&	356	&	226730	&	1255268	&	41.80	&	\textbf{0.787}	&	28.18	&	0.789	&	36.71	&	\textbf{0.787}	&	32.10	&	0.793	&	\textbf{6.93}	&	0.791	\\
    \midrule
    Overall & & & & 62.70 & 1.408 & 94.55 & 0.988 & 227.27 & \textbf{0.92} & 39.51 & 0.922 & \textbf{14.21} & 0.926\\
    \bottomrule
    \end{tabular}
    \label{table:ba:bal}
\end{table*}

\begin{table*}[t]
    \centering
    \renewcommand{\arraystretch}{1.2}
    \caption{Performance comparison with CPU-based BA frameworks on the 1DSfM dataset.}
    \begin{tabular}{L{0.1\linewidth}|C{0.044\linewidth} C{0.044\linewidth} C{0.044\linewidth}|C{0.041\linewidth} C{0.041\linewidth}|C{0.041\linewidth} C{0.041\linewidth}|C{0.041\linewidth} C{0.041\linewidth}|C{0.041\linewidth} C{0.041\linewidth}|C{0.041\linewidth} C{0.041\linewidth}}
    \toprule
    \multirow{2}{*}{Scene}	&	\multirow{2}{*}{Camera}	&	\multirow{2}{*}{Points}	&	\multirow{2}{*}{Pixels}	&	\multicolumn{2}{c|}{GTSAM \cite{gtsam}}	&	\multicolumn{2}{c|}{\gTwoO~\cite{g2o}}	&	\multicolumn{2}{c|}{Ceres \cite{ceres}}	&	\multicolumn{2}{c|}{\textbf{Ours} (PCG)}	&	\multicolumn{2}{c}{\textbf{Ours} (Cholesky)}		\\
     & & & & Time$\downarrow$	&	Error$\downarrow$	&	Time$\downarrow$	&	Error$\downarrow$	&	Time$\downarrow$	&	Error$\downarrow$	&	Time$\downarrow$	&	Error$\downarrow$	&	Time$\downarrow$	&	Error$\downarrow$	\\
    \midrule
    Union Square	&	166	&	3643	&	39651	&	9.53	&	2.365	&	0.78	&	2.617	&	2.15	&	\textbf{2.324}	&	1.21	&	2.358	&	\textbf{0.33}	&	2.377	\\
    P. del Popolo	&	317	&	13294	&	71055	&	10.79	&	3.104	&	5.64	&	3.104	&	8.42	&	3.102	&	1.61	&	\textbf{2.925}	&	\textbf{1.16}	&	2.928	\\
    Ellis Island	&	287	&	17565	&	64697	&	8.48	&	3.473	&	3.68	&	3.502	&	9.28	&	\textbf{3.446}	&	1.07	&	3.449	&	\textbf{0.60}	&	3.478	\\
    NYC Library	&	265	&	11247	&	50103	&	5.38	&	2.857	&	4.50	&	2.857	&	2.39	&	\textbf{2.855}	&	1.14	&	2.856	&	\textbf{0.53}	&	\textbf{2.855}	\\
    M. N. Dame	&	475	&	28209	&	147250	&	22.82	&	3.498	&	16.97	&	3.444	&	18.41	&	\textbf{3.426}	&	1.30	&	3.427	&	\textbf{1.25}	&	\textbf{3.426}	\\
    Gen. markt	&	745	&	32940	&	128472	&	10.82	&	4.793	&	45.45	&	2.968	&	30.66	&	\textbf{2.922}	&	1.62	&	2.925	&	\textbf{1.16}	&	2.928	\\
    Alamo	&	741	&	82801	&	536967	&	64.73	&	3.728	&	32.57	&	3.817	&	63.14	&	\textbf{3.726}	&	\textbf{3.25}	&	3.727	&	3.59	&	3.727	\\
    Yorkminster	&	64	&	3432	&	16351	&	2.70	&	2.244	&	\textbf{0.19}	&	2.323	&	1.67	&	\textbf{2.059}	&	2.43	&	2.090	&	{0.59}	&	2.094	\\
    Roman Forum	&	905	&	44245	&	151704	&	15.56	&	3.128	&	53.20	&	2.988	&	34.45	&	2.982	&	2.19	&	\textbf{2.980}	&	\textbf{0.97}	&	2.983	\\
    V. Cathedral	&	712	&	35688	&	170443	&	39.06	&	2.652	&	55.61	&	2.667	&	54.45	&	\textbf{2.634}	&	\textbf{1.90}	&	2.636	&	2.20	&	2.636	\\
    M. Metropolis	&	97	&	4981	&	21930	&	2.38	&	2.612	&	0.49	&	2.591	&	1.50	&	\textbf{2.588}	&	0.86	&	2.598	&	\textbf{0.29}	&	2.593	\\
    Piccadily	&	1898	&	83234	&	363139	&	233.57	&	3.737	&	454.24	&	3.484	&	290.14	&	3.419	&	\textbf{2.53}	&	\textbf{3.418}	&	13.71	&	3.423	\\
    T. of London	&	327	&	13156	&	58179	&	9.45	&	2.303	&	5.67	&	2.245	&	13.87	&	\textbf{2.098}	&	1.28	&	2.108	&	\textbf{0.77}	&	2.108	\\
    Trafalgar	&	4159	&	130027	&	572171	&	494.02	&	3.387	&	405.56	&	3.342	&	486.15	&	3.311	&	\textbf{2.96}	&	\textbf{3.241}	&	12.40	&	3.307	\\
    \midrule
    Overall & & & & 929.28	&	3.134	& 1084.55	&	2.996	& 1016.67	&	2.921   &	\textbf{25.35}    &	\textbf{2.910} &	39.55 &	2.919\\
    \bottomrule
    \end{tabular}
    \label{table:ba:1dsfm}
\end{table*}

\begin{table}[t]
	\centering
	\setlength{\tabcolsep}{0.10em}
	\renewcommand{\arraystretch}{1.2}
	\caption{Comparision with GPU-based methods.}
	\begin{tabular}{L{0.19\linewidth}|C{0.09\linewidth} C{0.09\linewidth}|C{0.09\linewidth} C{0.09\linewidth}|C{0.09\linewidth} C{0.09\linewidth}|C{0.09\linewidth} C{0.09\linewidth}}
	\toprule
	\multirow{2}{*}{Scene}	&	\multicolumn{2}{c|}{Ceres \cite{ceres}}	&	\multicolumn{2}{c|}{DeepLM \cite{huang2021deeplm}} &	\multicolumn{2}{c|}{Theseus \cite{pineda2022theseus}}	&	\multicolumn{2}{c}{\textbf{Ours} (Best)}	\\
	& Time$\downarrow$	&	Error$\downarrow$	& Time$\downarrow$	&	Error$\downarrow$	&	Time$\downarrow$	&	Error$\downarrow$	&	Time$\downarrow$	&	Error$\downarrow$	\\
	 \midrule
	Ladybug	&	56.90	&	1.144	&	5.87	&	1.121	&	606.34	&	1.581	&	\textbf{1.60}	&	\textbf{1.120}	\\
	Trafalgar	&	8.46	&	0.856	&	3.44	&	0.858	&	610.18	&	9.149	&	\textbf{1.27}	&	\textbf{0.853}	\\
	Dubrovnik	&	19.00	&	\textbf{0.786}	&	13.10	&	0.787	&	2597.1	&	11.649	&	\textbf{6.93}	&	0.791	\\
	\midrule
	BAL Overall &	84.36	&	0.929	& 22.41 & 0.922 & 3813.6	&	7.460	&	\textbf{9.80} & \textbf{0.921}\\
	\midrule
	\scriptsize{Union Square}	&	3.42	&	\textbf{2.326}	&	1.31	&	2.330	&	59.57	&	2.585	&	\textbf{0.33}	&	2.377	\\
	P. del Popolo	&	4.77	&	3.102	&	1.45	&	3.103	&	95.48	&	3.398	&	\textbf{1.16}	&	\textbf{2.928}	\\
	Ellis Island	&	6.24	&	\textbf{3.446}	&	1.46	&	3.448	&	93.95	&	3.652	&	\textbf{0.60}	&	3.478	\\
	NYC Library	&	1.90	&	\textbf{2.854}	&	1.40	&	2.855	&	72.33	&	3.058	&	\textbf{0.53}	&	2.855	\\
	M. N. Dame	&	14.79	&	\textbf{3.426}	&	1.76	&	\textbf{3.426}	&	204.00	&	3.607	&	\textbf{1.25}	&	\textbf{3.426}	\\
	Gen. markt	&	15.23	&	\textbf{2.922}	&	1.77	&	2.926	&	167.94	&	3.200	&	\textbf{1.16}	&	2.928	\\
	Alamo	&	133.98	&	\textbf{3.726}	&	3.43	&	3.727	&	848.18	&	3.864	&	\textbf{3.25}	&	3.727	\\
	Yorkminster	&	0.79	&	\textbf{2.058}	&	1.24	&	2.089	&	25.06	&	2.424	&	\textbf{0.59}	&	2.094	\\
	\scriptsize{Roman Forum}	&	21.99	&	\textbf{2.978}	&	1.65	&	2.984	&	236.89	&	3.269	&	\textbf{0.97}	&	2.983	\\
	V. Cathedral	&	24.57	&	\textbf{2.634}	&	2.04	&	2.636	&	220.48	&	3.086	&	\textbf{1.90}	&	2.636	\\
	\scriptsize{M. Metropolis}	&	0.72	&	\textbf{2.588}	&	1.20	&	2.589	&	34.06	&	2.792	&	\textbf{0.29}	&	2.593	\\
	Piccadily	&	240.0	&	\textbf{3.416}	&	\textbf{2.28}	&	3.419	&	478.05	&	3.600	&	{2.53}	&	3.418	\\
	T. of London	&	4.66	&	\textbf{2.100}	&	1.42	&	2.103	&	83.05	&	2.407	&	\textbf{0.77}	&	2.108	\\
	Trafalgar	&	1078.5	&	\textbf{3.238}	&	3.05	&	3.241	&	1130.1	&	3.450	&	\textbf{2.96}	&	3.241	\\
	\midrule
	\scriptsize{1DSfM} Overall &	1551.6	&	2.915	& 25.45 & 2.92 & 3749.1	&	3.171	&	\textbf{18.29} & \textbf{2.913}\\
	\bottomrule
	\end{tabular}
	\vspace{-5pt}
	\label{table:deeplm}
	\end{table}

\section{Experiments}

We next conduct extensive experiments to compare our BA in the eager mode with the popular BA libraries.

\label{exp:ba}
\subsection{Datasets, Baseline, Platforms, and Metrics}
\label{exp:ba:setup}
We conduct experiments on three datasets: \textbf{BAL}, \textbf{1DSfM}, and \textbf{CO3D v2}. The BAL \cite{bal} and 1DSfM \cite{wilson_eccv2014_1dsfm} datasets are used for benchmark evaluation. The BAL dataset provides the initial estimates of 3D maps and camera locations, whereas 1DSfM provides only raw images of the wild in Internet photo collections.
Following \cite{huang2021deeplm}, we generate the initial map for 1DSfM using COLMAP with its BA disabled. The CO3D v2 \cite{reizenstein21common} is used to assess our BA framework in a realistic deep learning pipeline in \sref{sec:vggt}.

We conduct experiments using our PCG and Cholesky sparse linear solvers, denoted as Ours (PCG) and Ours (Cholesky).
They are compared against the most widely-used BA frameworks, including Ceres Solver \cite{ceres}, \gTwoO \cite{g2o}, and GTSAM \cite{gtsam}.
Ceres Solver is widely regarded as the leading BA library, known for its robustness and scalability to efficiently leverage a large number of CPU cores.
Additionally, to ensure their best efficiency on CPU, we compiled GTSAM with Intel OneTBB \cite{intel_oneTBB}, and \gTwoO\ and Ceres Solver were built using OpenMP \cite{openmp08}, with an optimization flag ``-O3" applied. All CPU-based libraries are evaluated on a high-end dual-socket server with 2 $\times$ AMD EPYC 7543 (32 cores each) and 512 GB of memory.
We also compare with the GPU-based framework DeepLM \cite{huang2021deeplm}, Theseus \cite{pineda2022theseus}, and Ceres Solver \cite{ceres}. The performance is presented using an Nvidia RTX 4090 GPU with double-precision floating point arithmetic.

For evaluation, we assess the frameworks based on two metrics on BAL and 1DSfM: reprojection mean squared error (MSE) in pixels to measure accuracy, and runtime in seconds to evaluate runtime efficiency. The reported runtime measures only the optimization stage after the input problem has been instantiated. It excludes one-time dataset parsing, e.g., loading camera variables, point variables, and observation correspondences. This focuses the comparison on solver performance and convergence quality rather than data-loading or front-end setup overhead. The benchmark follows each library's official user-facing implementation: DeepLM, Theseus, and ours are implemented in Python and PyTorch, while GTSAM, Ceres Solver, and \gTwoO\ are evaluated through their C++ interfaces.

\subsection{Overall Performance}

\subsubsection{BAL Dataset}
\begin{figure}[ht]
  \centering
\includegraphics[width=0.49\linewidth]{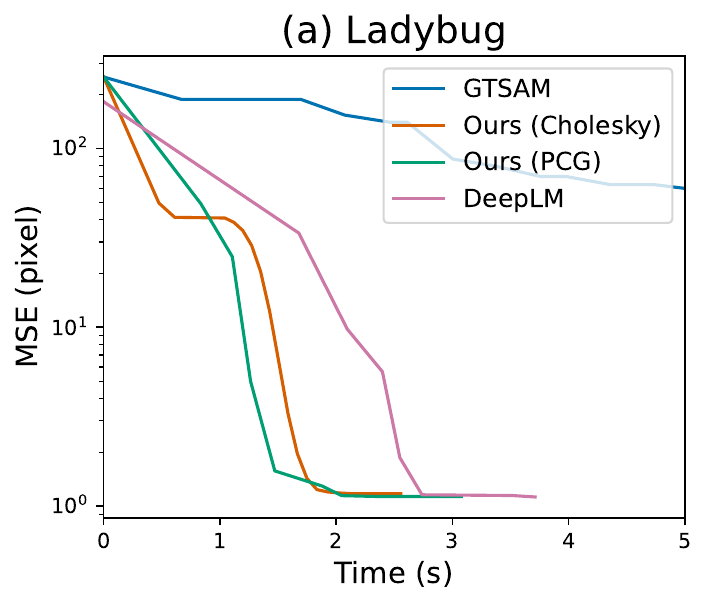}
  \hfill
\includegraphics[width=0.49\linewidth]{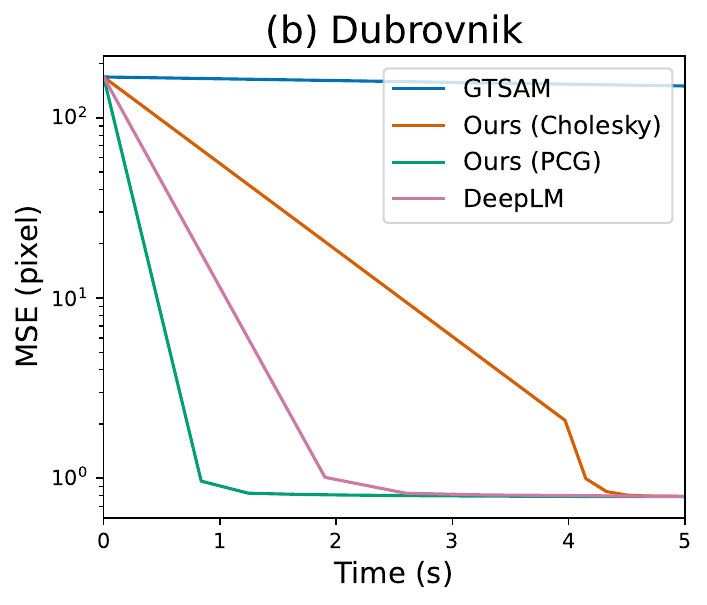}

    \includegraphics[width=0.49\linewidth]{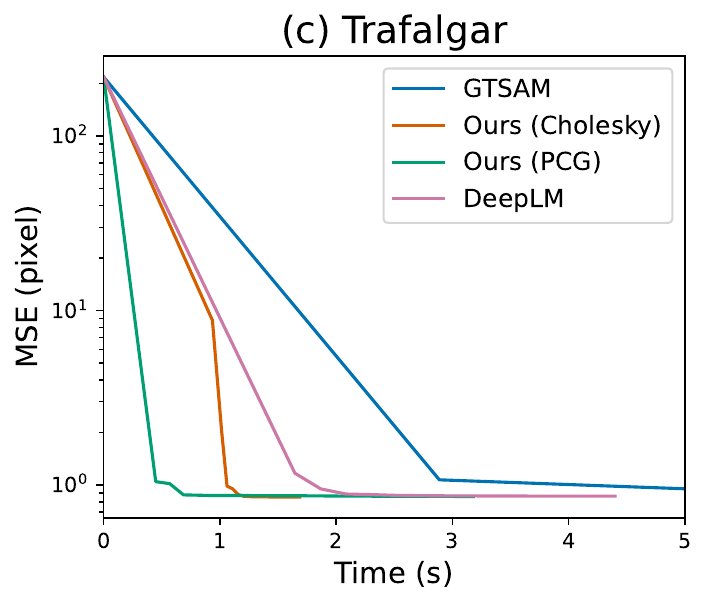}
  \hfill
    \includegraphics[width=0.49\linewidth]{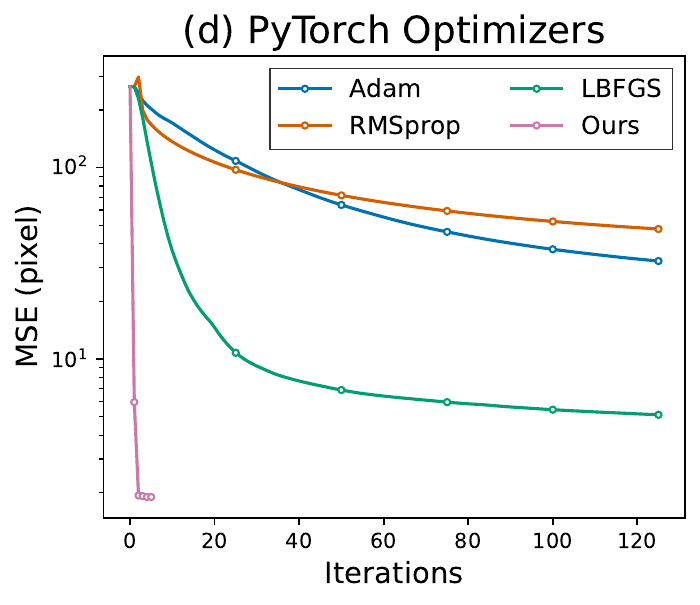}
  \caption{(a)-(c): Convergence curves (MSE v.s. Time) on the BAL dataset. (d): Comparison with native PyTorch optimizers on Ladybug.}
  \label{fig:convergence_curve}
\end{figure}

\begin{figure*}[t]
    \centering
     \includegraphics[width=\textwidth]{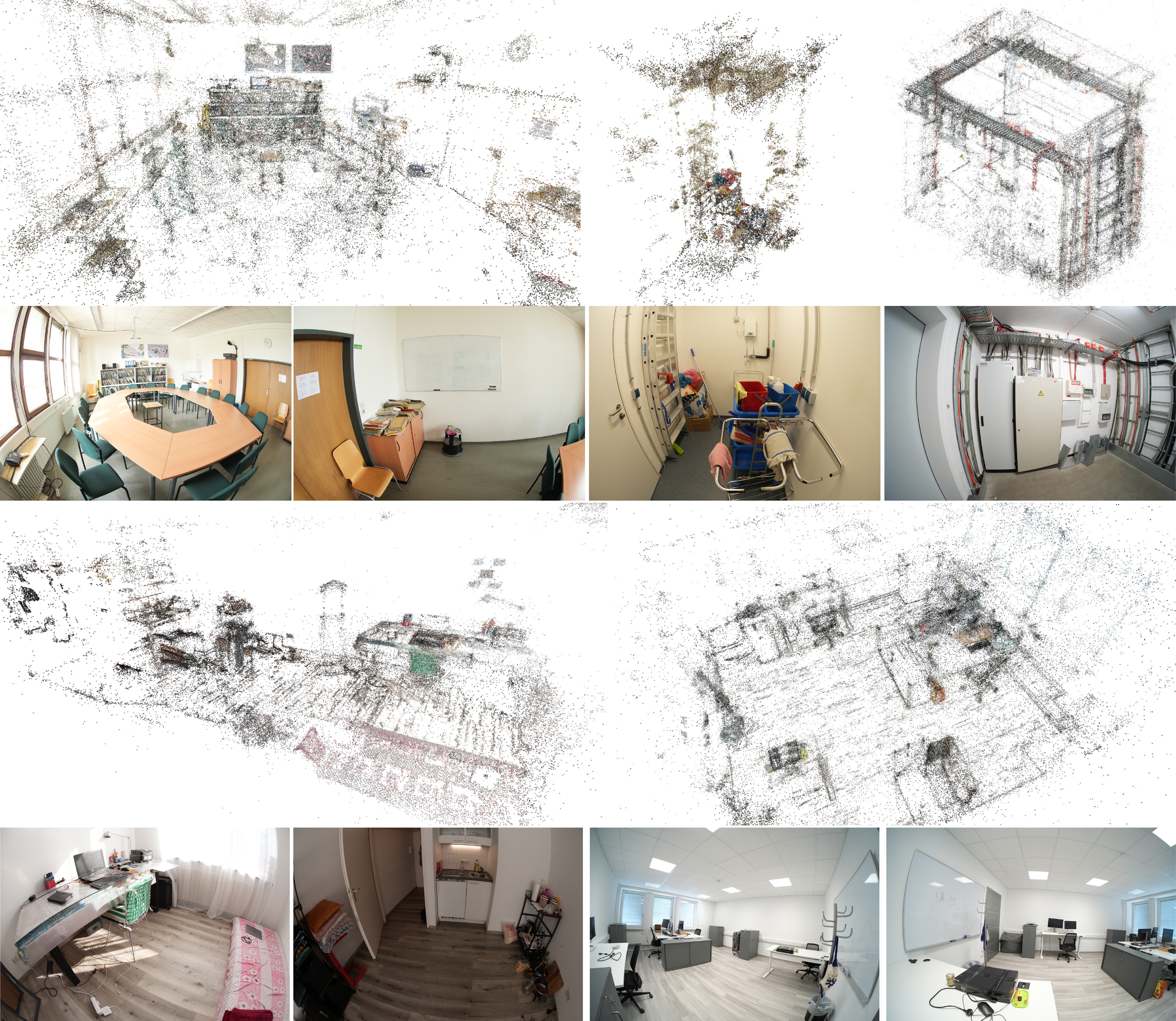}
     \caption{
     Qualitative 3D reconstruction results on diverse indoor environments \cite{yeshwanth2023scannet++}. Top rows: point cloud reconstructions showcasing structural clarity and spatial accuracy. Bottom rows: corresponding reference images capturing varied indoor settings including office rooms, storage areas, and residential spaces.}
     \label{fig:scannet}
\end{figure*}

The performance comparison on the BAL dataset is presented in \tref{table:ba:bal}. Our BA in the eager mode achieves much higher efficiency, i.e., 4.4$\times$, 6.7$\times$, and 16$\times$ faster than GTSAM, \gTwoO, and Ceres Solver, respectively.
It is observed that all the BA frameworks except GTSAM can converge. Therefore, their precision in terms of MSE is comparable.
We also noticed that Ours (Cholesky) achieves higher efficiency than Ours (PCG) in the scenes of ``Trafalgar" and ``Dubrovnik''. This is because ``Trafalgar" has fewer parameters so that the direct linear solver can quickly perform pivoting, and ``Dubrovnik'' has an ill-posed linear system, causing PCG more iterations to converge. The qualitative results on the three scenarios are shown in \fref{fig:ba_BAL}, showing a high level of detail reconstructed. Additional in-the-wild qualitative samples are provided in \fref{fig:scannet}. \fref{fig:convergence_curve} shows the convergence time curve, and our method requires the shortest time compared with the rest of the baselines. Note that none of the PyTorch \nth{1}-order optimizers is able to converge. 

\subsubsection{1DSfM Dataset}
We present the overall performance on the 1DSfM dataset in \tref{table:ba:1dsfm}, where all frameworks share a similar final error.
In terms of runtime efficiency, our BA framework surpasses GTSAM by {36$\times$, \gTwoO\ by 43$\times$}, and Ceres Solver by 40$\times$ in running speed, further demonstrating a consistently high efficiency of our BA in the eager mode. 

\begin{figure}[t]
    \centering
     \includegraphics[width=1.0\linewidth]{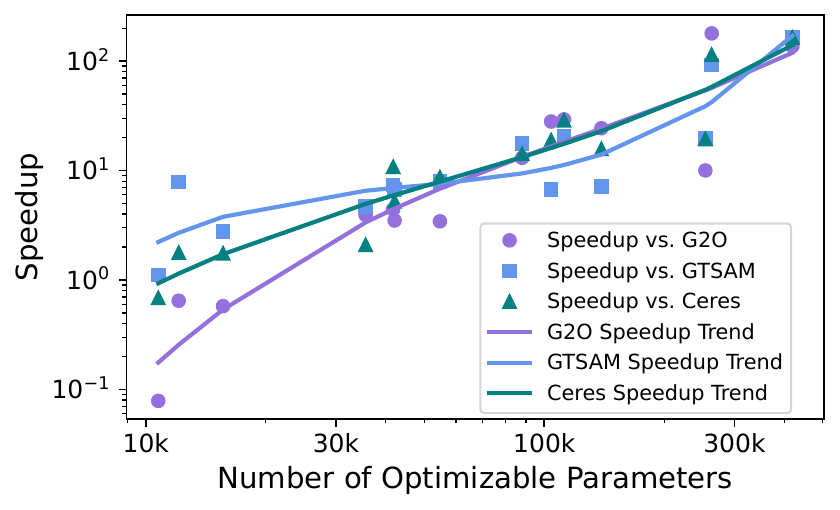}
     \caption{Speedup of our BA relative to other frameworks exponentially increases with the number of optimizable parameters.}
     \label{fig:ba_scaling}
\end{figure}

\subsubsection{Scalability}

We next demonstrate the scalability of our BA in terms of the number of optimizable parameters. 
\fref{fig:ba_scaling} illustrates the runtime speedup of our PCG optimizer relative to \gTwoO, GTSAM, and Ceres Solver on the 1DSfM data samples. 
The plot reveals a general trend: as the problem scale increases, our BA demonstrates \textbf{exponentially} increased efficiency, reaching up to 136$\times$, 166$\times$, and 163$\times$ higher efficiency compared to \gTwoO, GTSAM, and Ceres Solver, respectively.
On small samples, our performance is bounded by the Python interpreter overhead and thus is similar to other libraries. Despite eager mode's performance limitations without compile-time optimization, the significant speedups result from our sparsity-aware algorithm, which efficiently leverages inherent sparsity and enables effective GPU parallelism.

\begin{table}[t]
    \centering
    \caption{Comparison with Deep Learning SfM Pipeline on CO3Dv2}
    \resizebox{0.90\columnwidth}{!}{
    \begin{tabular}{l|c|c}
    \toprule
    Methods & CO3Dv2 & Time \\
             & AUC@30 $\uparrow$ & \\
    \midrule
    COLMAP+SPSG~\cite{sarlin2020superglue}    & 25.3              &  $\sim$ 15s          \\
    PixSfM~\cite{lindenberger21pixel-perfect} & 30.1              &  $>$ 20s             \\
    PoseDiff~\cite{wang23posediffusion:}      & 66.5              &  $\sim$ 7s           \\
    DUSt3R~\cite{wang24dust3r:}               & 76.7              &  $\sim$ 7s           \\
    MASt3R~\cite{mast3r}                      & 81.8              &  $\sim$ 9s           \\
    VGGSfM v2~\cite{wang24vggsfm:}            & 83.4              &  $\sim$ 10s          \\ 
    MV-DUSt3R~\cite{tang2024mv} & 69.5 & {$\sim$ 0.6s} \\
    CUT3R~\cite{cut3r} & 82.8 & {$\sim$ 0.6s}  \\
    FLARE~\cite{zhang2025flare} & 83.3 & \underline{{$\sim$ 0.5s}}  \\
    Fast3R~\cite{yang2025fast3r}  & 82.5 & \textbf{$\sim$ 0.2s}  \\
    \midrule
    VGGT (Initialization only) \cite{wang2025vggt}                       & \underline{88.2}  & \textbf{$\sim$ 0.2s} \\
    VGGT + PyCOLMAP \cite{wang2025vggt}                           & \textbf{90.0}     &  $\sim$ 1.8s         \\
    \textbf{VGGT + Ours}                            & \textbf{90.0}     &  $\sim$ 0.9s         \\
    \bottomrule
    \end{tabular}
    }
    \label{tab:vggt}
\end{table}

\begin{figure}[t]
    \centering
     \includegraphics[width=1.0\linewidth]{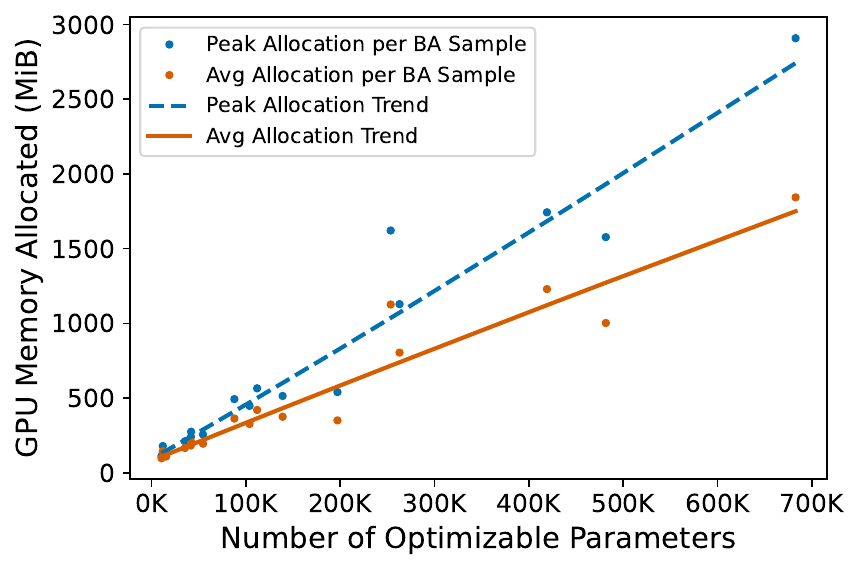}
     \caption{\textbf{Memory Consumption Trend}. Memory consumption increases linearly with the number of parameters as it grows larger than 30k. Peak allocation is consistently higher than average due to sparse linear algebra operations shortly using temporary buffers.}
     \label{fig:memory}
\end{figure}

\subsubsection{GPU Memory Usage}

We measure GPU memory consumption for BAL scenes and report average memory usage and peak CUDA memory allocation in \tref{tab:gpu_memory}. The overall trend for all BAL and 1DSfM scenarios is shown in \fref{fig:memory}. 
These memory usage are reported using the PCG solver, and peak and average memory statistics are provided by the PyTorch CUDA memory profiler. 
The PCG solver is entirely implemented with PyTorch operators, so all its allocations are captured by PyTorch memory profiler. 
In contrast, Cholesky relies on vendor linear algebra libraries that use working buffers not attributed in PyTorch’s memory profiling, so the estimated numbers can be inaccurate for GPU memory.
Our method exhibits linear predictable memory usage, well below the capacity of modern commodity GPUs, leaving ample headroom for integration with deep learning models. The gap between peak and average memory usage reflects transient spikes primarily caused by temporary work buffers generated during sparse operations such as matrix transposition.
These temporary usages dominate peak allocation but are released immediately after each LM iteration is initialized.

\begin{table}[t]
\centering
\caption{GPU memory usage for BAL scenes.}
\label{tab:gpu_memory}
\begin{tabular}{lcc}
\toprule
\textbf{Scene} & \textbf{Peak CUDA Memory} & \textbf{Average CUDA Memory} \\
\midrule
Dubrovnik  & 2.84~GiB & 1.8~GiB \\
Ladybug   & 1.54~GiB & 1002.4~MiB \\
Trafalgar & 539.75~MiB & 350.4~MiB \\
\bottomrule
\end{tabular}
\end{table}

\subsubsection{Comparison with GPU-based framework}

As shown in \tref{table:deeplm}, our BA framework achieves both higher precision and efficiency compared to other GPU-supported solutions, including Ceres Solver \cite{ceres}, DeepLM \cite{huang2021deeplm}, and Theseus \cite{pineda2022theseus}. Ceres Solver \cite{ceres} is compiled with its optional CUDA support. DeepLM \cite{huang2021deeplm} is the state-of-the-art PyTorch-based BA solution. Theseus \cite{pineda2022theseus} uses PyTorch as the foundation but extends it with CUDA implementations to handle sparse linear algebra and batched operations. We compile Theseus with BaSpaCho \cite{facebookresearch_baspacho}, its designated sparse CUDA linear solver. 

Our method is the fastest among these GPU-supported solutions. Specifically, we require 56\% and 28\% less runtime than DeepLM on the BAL and 1DSfM datasets, respectively. Theseus fails to converge on the \texttt{Ladybug-1723} and \texttt{Trafalgar-257} samples in BAL and reports numerical issues. We instead report its accuracy on the largest possible samples it can solve, \texttt{Ladybug-539} and \texttt{Trafalgar-201} from the same scenes.
Note that although DeepLM and Theseus are based on PyTorch, they do not support PyTorch eager mode and lack extensibility to other applications, since their sparsity is addressed by customized non-native data structures.

\begin{figure*}[t]
    \centering
     \includegraphics[width=\textwidth]{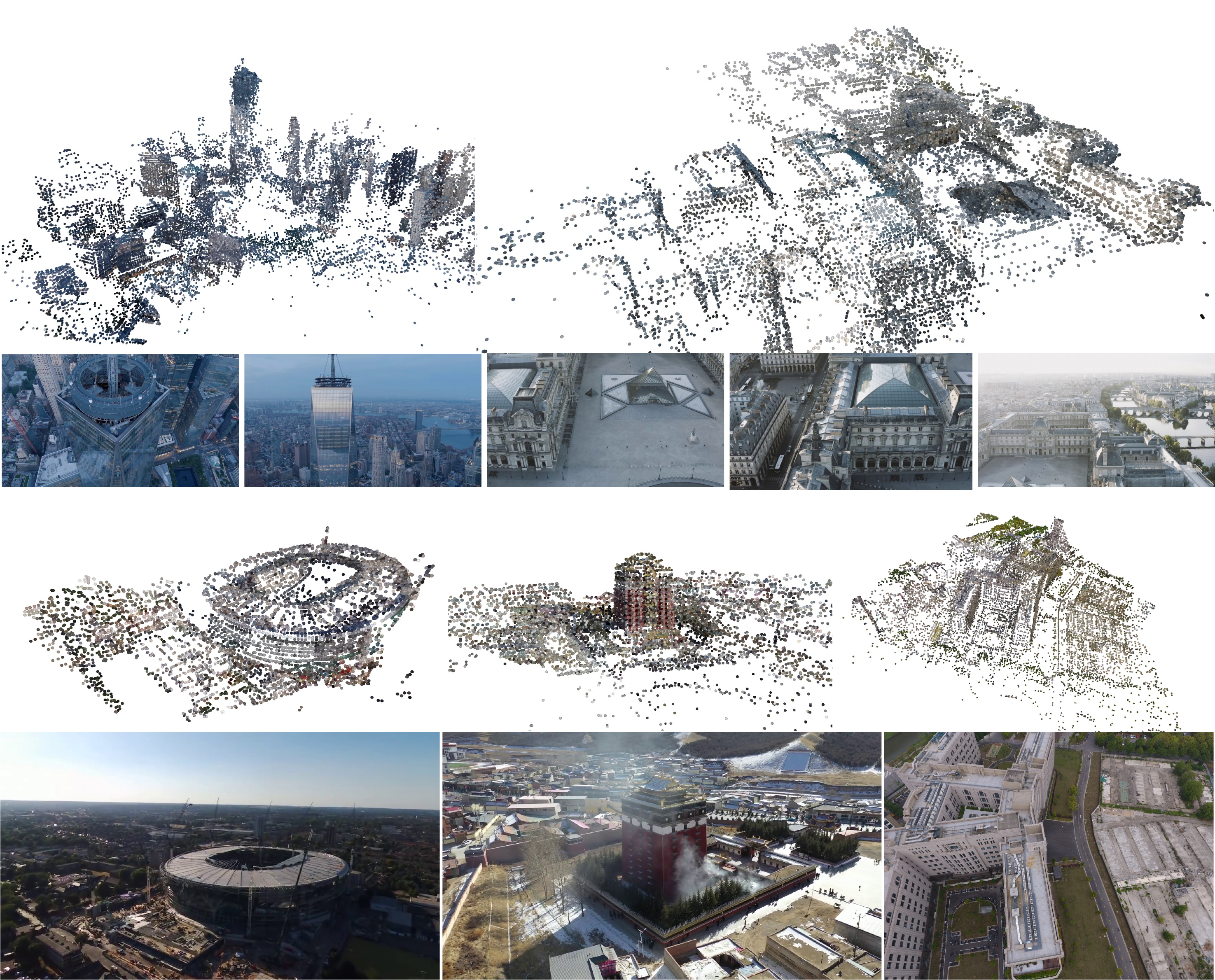}
     \caption{
     Qualitative 3D reconstruction results on outdoor scenes \cite{lu2023ommo}, demonstrating accurate geometric initialization and refinement with deep-learning-based SfM pipeline, VGGT \cite{wang2025vggt}. Top rows: point cloud reconstructions of urban landmarks capturing detailed architectural and structural elements. Bottom rows: corresponding aerial reference images illustrating diverse urban scenes, including skyscrapers, stadiums, historical monuments, and complex cityscapes. }
     \label{fig:ommo}
\end{figure*}

\begin{figure*}[ht]
  \centering
\includegraphics[width=0.2455\linewidth]{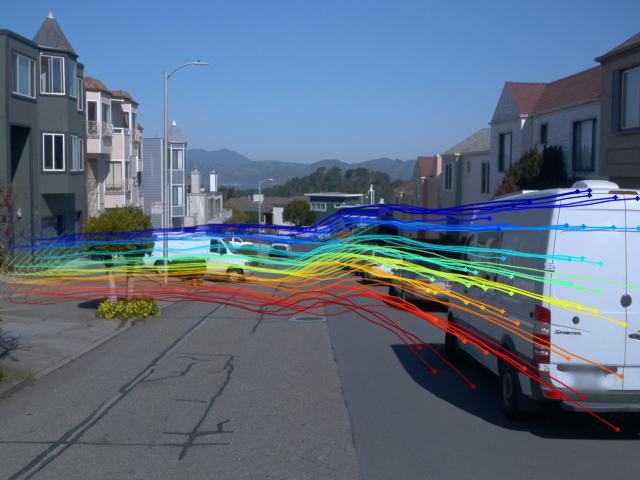}
\includegraphics[width=0.2455\linewidth]{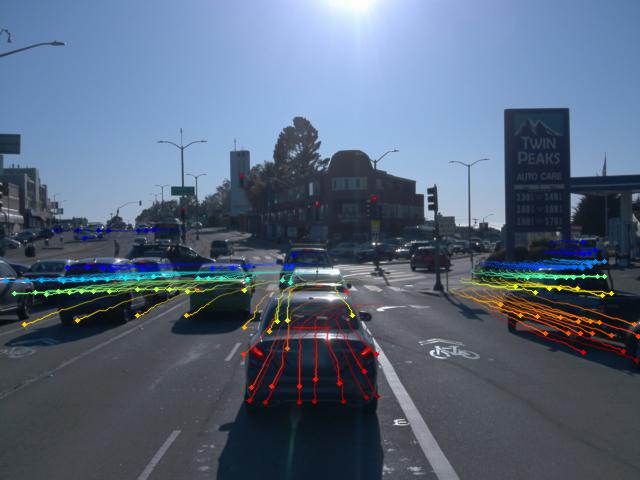}
    \includegraphics[width=0.2455\linewidth]{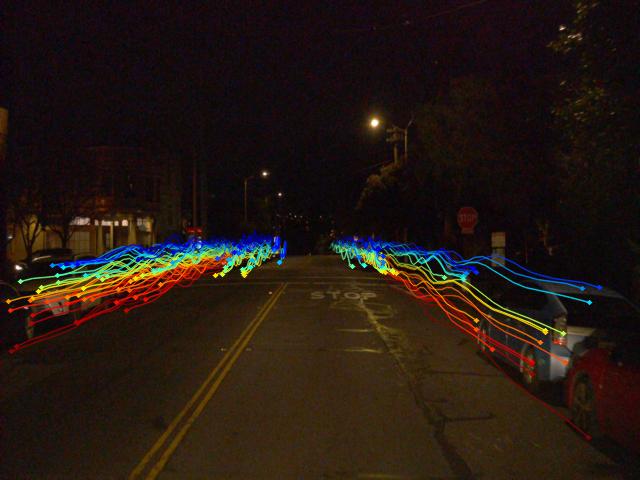}
    \includegraphics[width=0.2455\linewidth]{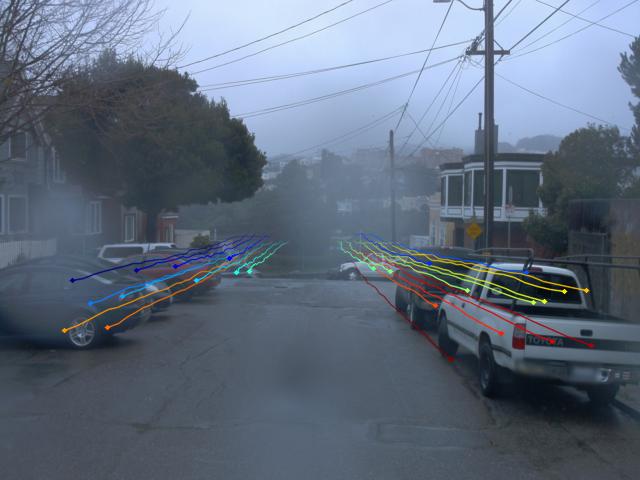}
    
\vspace{5pt}
    \includegraphics[width=0.2455\linewidth]{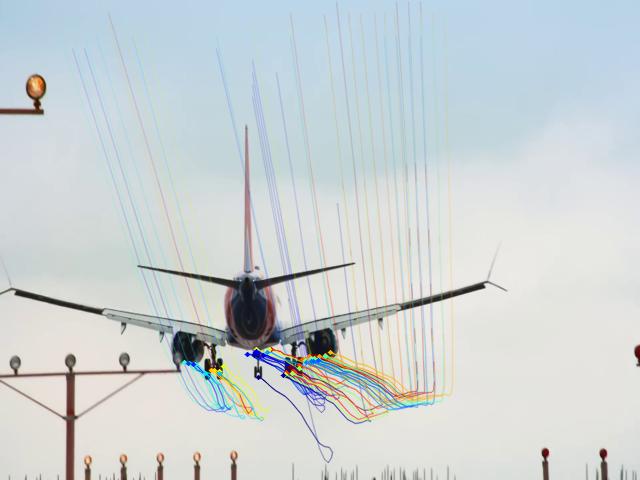}
    \includegraphics[width=0.2455\linewidth]{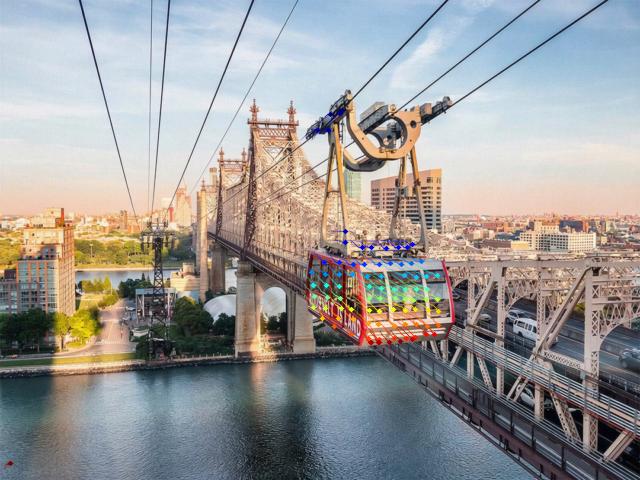}
    \includegraphics[width=0.2455\linewidth]{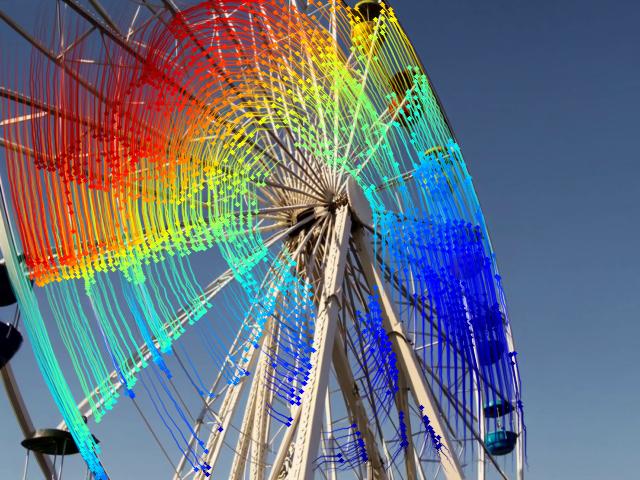}
    \includegraphics[width=0.2455\linewidth]{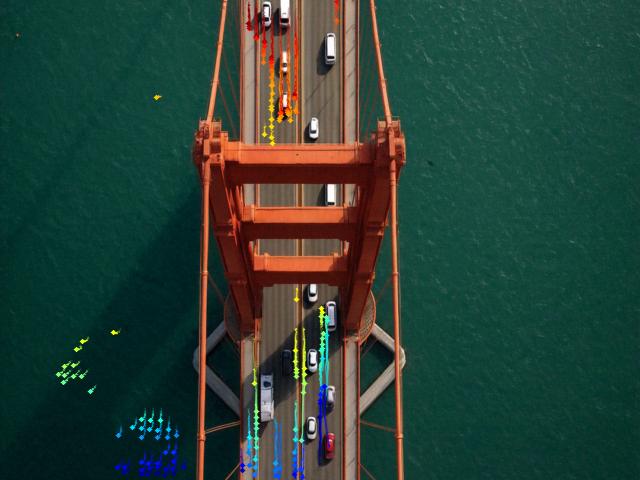}
    
\vspace{5pt}
    \includegraphics[width=0.2455\linewidth]{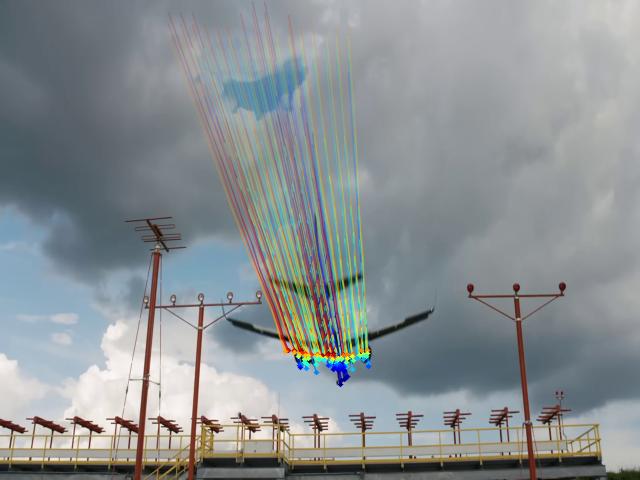}
    \includegraphics[width=0.2455\linewidth]{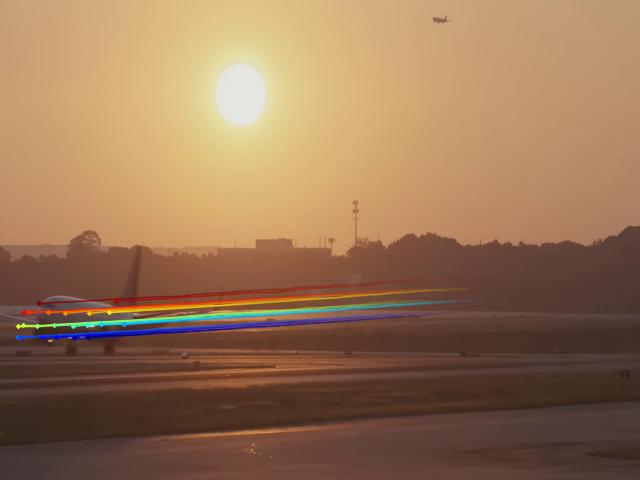}
    \includegraphics[width=0.2455\linewidth]{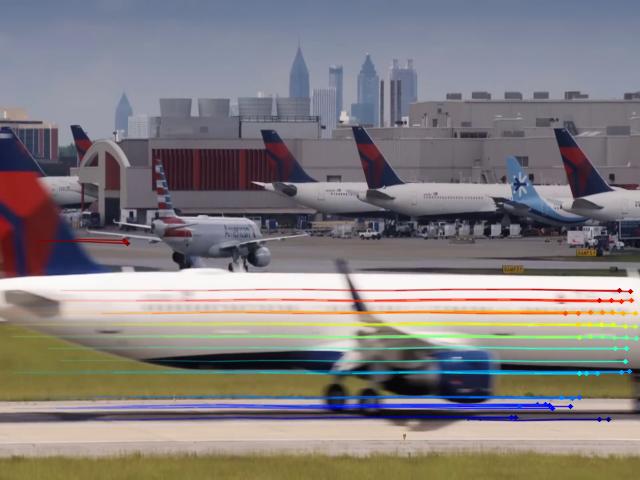}
    \includegraphics[width=0.2455\linewidth]{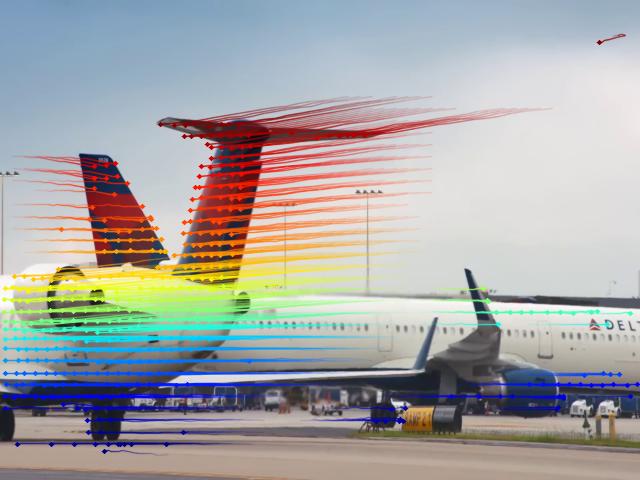}
    
\vspace{5pt}
    \includegraphics[width=0.2455\linewidth]{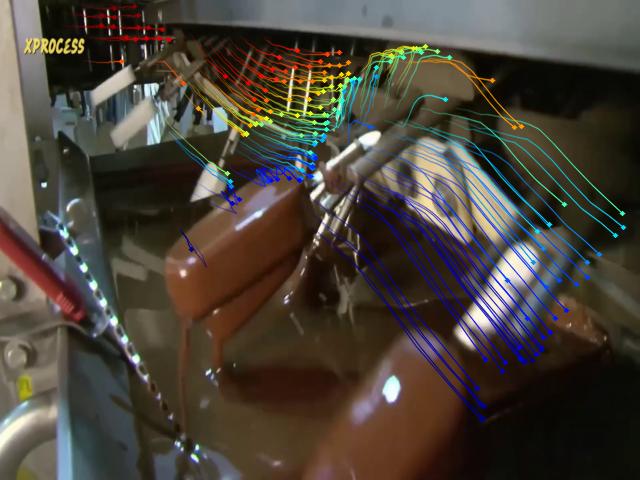}
    \includegraphics[width=0.2455\linewidth]{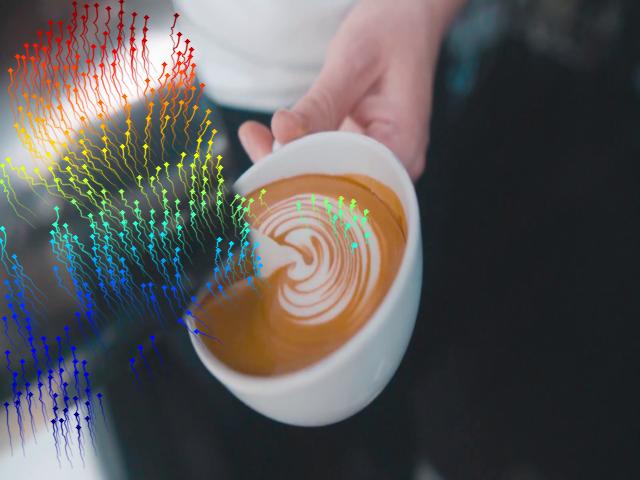}
    \includegraphics[width=0.2455\linewidth]{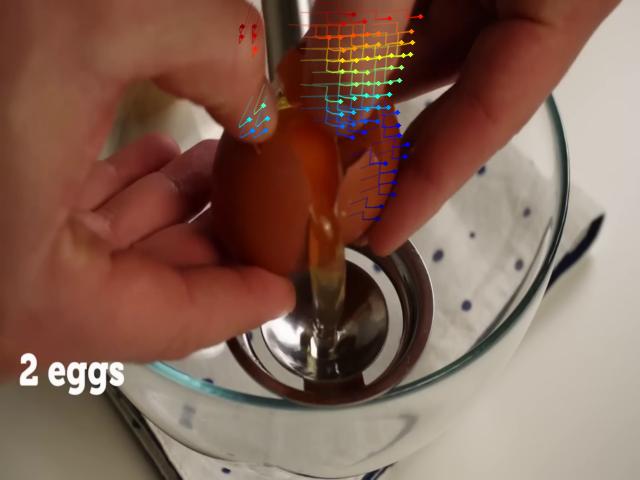}
    \includegraphics[width=0.2455\linewidth]{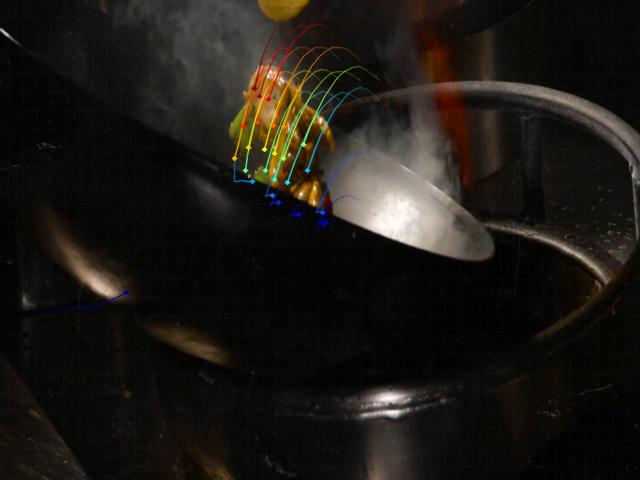}
  \caption{Zero-shot qualitative evaluation of the iDKM model on the Waymo Open Dataset (row 1) and diverse real-world scenarios (rows 2-4). Our approach showcases robust generalization capabilities, effectively capturing intricate motion patterns across various dynamic contexts, including driving, aviation, and everyday activities, through self-supervised learning.}
  \label{fig:waymo_dkm}
\end{figure*}

\subsection{Integration with Deep Learning SfM Pipeline}
\label{sec:vggt}

To further validate the flexibility and performance advantages of our library, we integrated it with the Visual Geometry Grounded Transformer (VGGT)~\cite{wang2025vggt}, a state-of-the-art deep-learning-based SfM method. VGGT leverages a large-scale transformer to predict key 3D attributes of a scene, including camera extrinsics and intrinsics, depth maps, 3D point clouds, and point correspondences. While VGGT provides rapid initialization within less than a second for hundreds of views, its original implementation relies on PyCOLMAP for post-processing BA, leading to bottlenecks due to limited CPU throughput and costly data transfers between CPU and GPU. By integrating VGGT predictions directly with our BA library, we enabled the \textit{entire} SfM pipeline to operate efficiently in PyTorch eager mode, eliminating these data transfer overheads and significantly reducing latency. 

We evaluated on the CO3D v2 dataset~\cite{reizenstein21common}, a large-scale collection of multi-view videos across diverse object categories. The evaluation focuses on camera pose estimation accuracy, using the Area-under-Curve (AUC) metric at a 30-degree threshold, denoted as AUC@30, which is widely adopted for benchmarking deep-learning-based SfM pipelines \cite{wang2025vggt}. In addition to VGGT, we include comparisons with a comprehensive set of recent baselines, including COLMAP with SuperGlue features~\cite{sarlin2020superglue} denoted as \texttt{COLMAP+SPSG}, PixSfM~\cite{lindenberger21pixel-perfect}, PoseDiff~\cite{wang23posediffusion:}, DUSt3R~\cite{wang24dust3r:}, MASt3R~\cite{mast3r}, and VGGSfM v2~\cite{wang24vggsfm:}, as well as fast multi-view variants like MV-DUSt3R~\cite{tang2024mv}, CUT3R~\cite{cut3r}, FLARE~\cite{zhang2025flare}, and Fast3R~\cite{yang2025fast3r}. For fair comparison, we report runtime and accuracy under the same GPU environment as \cite{wang2025vggt}.

Our experiments summarized in \tref{tab:vggt} demonstrate that the accuracy of our integrated approach remains consistent with the original VGGT implementation, confirming its correctness and reliability. Importantly, our approach reduces the BA runtime substantially, achieving an average optimization time of \textbf{0.7s}, which is 2.3$\times$ faster than PyCOLMAP. This makes our method the \textbf{fastest} pipeline achieving SOTA accuracy. The efficiency gain primarily arises from seamless integration with PyTorch eager mode without costly CPU-GPU data transfers. Our results not only highlight clear latency improvements of our method but also demonstrate the ease with which our method can be incorporated into existing deep learning SfM pipelines, without the need for extensive engineering efforts. Additional qualitative evaluations on in-the-wild samples are provided in \fref{fig:ommo}.

\subsection{Integration with Self-supervised Feature Matching}

To show the versatility and integration capabilities of our BA framework, we applied it within the context of the iMatching \cite{zhan2024imatching} methodology for self-supervised feature correspondence learning. A description of iMatching is provided in Appendix \ref{x:imatching}.
We aimed to demonstrate its effectiveness as a plug-and-play replacement for traditional BA solvers within sophisticated learning frameworks, potentially simplifying the development cycle due to its native PyTorch interface. 

While the original iMatching paper proposes an efficient method for gradient backpropagation through BA leveraging stationary points, its practical implementations rely on GTSAM \cite{gtsam}. GTSAM operates outside the PyTorch ecosystem, requiring additional interfaces to handle gradient computations and data transfer between C++ and Python. This increases code complexity especially when implementing high-frequency training loops.

\begin{table}[t]
    \centering
    \caption{Learning Feature Matching  on KITTI360}
    \resizebox{\columnwidth}{!}    {
    \begin{tabular}{c|ccc|c|ccc}
        \toprule
        Method & $5^\circ$ & $10^\circ$ & $20^\circ$ & Method & $5^\circ$ & $10^\circ$ & $20^\circ$   \\
        \midrule
        ASpanFormer & 74.4 & 86.3 & 93.0  & DKM & 91.8 & 95.9 & 98.0 \\
        {iASpan \cite{zhan2024imatching}} & {80.6} & {90.1} & {95.1} & {iDKM \cite{zhan2024imatching}} & \textbf{92.6} & \textbf{96.3} & \textbf{98.1}  \\
        \textbf{iASpan (Ours)} & \textbf{83.2} & \textbf{91.5} & \textbf{95.7} & \textbf{iDKM (Ours)} & \textbf{92.6} & \textbf{96.3} & \textbf{98.1}  \\
        \bottomrule
    \end{tabular}
    }
    \label{tab:kitti}

\end{table}

In this experiment, we replaced the conventional BA component with our proposed framework. This integration ensures that gradients of self-supervision loss are computed efficiently within the same computational graph as the neural network, streamlining the training process.

\subsubsection{Datasets}
We report our results on the KITTI360 \cite{Liao2022PAMI} and ETH3D-SLAM \cite{eth3d} datasets. KITTI360 is a large-scale outdoor driving dataset, which contains 83k frames recording 73km of scene footage. It also introduces real-world challenges including dynamic objects (e.g., moving vehicles and pedestrians) and repetitive structures (e.g., buildings or road markings), all of which test the robustness and adaptability of feature matching models. 
Following \cite{zhan2024imatching}, we divide the entire KITTI360 dataset into subsets of size 8.5:0.5:1 for training, validation, and testing, respectively. 

The ETH3D-SLAM dataset \cite{eth3d}, in contrast, is an indoor SLAM dataset comprising small-scale scenes with high-precision ground truth. Its controlled environments and diverse camera motions make it ideal for evaluating the precision of our BA framework in scenarios with intricate geometry and limited spatial extent. We adopted the training and testing sequences defined in \cite{zhan2024imatching} ensuring consistency with prior work.

\subsubsection{Evaluation Metrics}
All feature matching models are evaluated using the widely-adopted pose estimation task and the Area Under the Curve (AUC) metric. Specifically, pose estimation accuracy is measured by computing the AUC for rotation and translation errors at thresholds of $5^\circ$, $10^\circ$, and $20^\circ$. For ground truth camera poses, the outdoor KITTI360 dataset leverages onboard sensors, including stereo cameras, LiDAR, and GPS, while the indoor ETH3D-SLAM dataset uses precise camera poses acquired through motion capture.

\subsubsection{Baselines}
Our experiments involve three state-of-the-art baseline models studied in the original iMatching paper: CAPS \cite{wang2020caps} employs an expectation-based approach \cite{zhan2024imatching} for differentiable sparse feature matching; ASpanFormer \cite{chen2022aspanformer} and DKM \cite{edstedt2023dkm} represent dense matching that leverage regression-based predictions \cite{zhan2024imatching}. 
We name our finetuned models as iCAPS, iASpan, and iDKM, respectively, and compare with their pretrained counterparts to examine whether they could successfully adapt to new testing scenes.

\begin{table*}[t]
  \centering
  \small
  \setlength{\tabcolsep}{0.26em}
  \caption{Self-supervised feature matching on ETH3D-SLAM.}
  \renewcommand{\arraystretch}{1.5}
    \resizebox{0.9\textwidth}{!}{
      \begin{tabular}{c|ccc|ccc!{\color{lightgray}\vrule}ccc!{\color{lightgray}\vrule}cccc|ccc!{\color{lightgray}\vrule}cccc}
      \toprule
      
      Method & \multicolumn{3}{c|}{SuperGlue~\cite{sarlin2020superglue}} & \multicolumn{3}{c}{SGP~\cite{yang2021self}} & \multicolumn{3}{c}{CAPS~\cite{wang2020caps}} & \multicolumn{4}{c|}{\textbf{iCAPS (Ours)}} & \multicolumn{3}{c}{DKM~\cite{edstedt2023dkm}} & \multicolumn{4}{c}{\textbf{iDKM (Ours)}} \\
      \midrule
      AUC & 5\textdegree & 10\textdegree & 20\textdegree  & 5\textdegree & 10\textdegree & 20\textdegree  & 5\textdegree & 10\textdegree & 20\textdegree  & 5\textdegree & 10\textdegree & 20\textdegree & \% $\uparrow$ & 5\textdegree & 10\textdegree & 20\textdegree  & 5\textdegree & 10\textdegree & 20\textdegree & \% $\uparrow$ \\
      \midrule
      cables & 66.9 & 72.6 & 75.4 & 62.0 & 67.9 & 70.9 & 67.4 & 73.9 & 77.1 & 70.3 & 77.6 & 81.2 & 4.30\% & 59.2 & 63.7 & 66.0 & 71.9 & 77.7 & 80.7 & 21.45\% \\
      camera\_shake & 64.5 & 78.0 & 86.1 & 68.4 & 82.1 & 89.8 & 63.0 & 77.8 & 87.0 & 71.3 & 83.7 & 91.2 & 13.17\% & 65.9 & 72.9 & 76.5 & 73.2 & 81.0 & 84.9 & 11.08\% \\
      ceiling & 81.0 & 86.6 & 89.7 & 78.9 & 85.0 & 88.1 & 81.4 & 87.8 & 91.1 & 83.3 & 89.9 & 93.3 & 2.33\% & 59.4 & 62.8 & 64.6 & 82.8 & 88.9 & 92.2 & 39.39\% \\
      desk & 77.1 & 84.5 & 88.4 & 73.5 & 82.3 & 87.0 & 72.8 & 82.2 & 87.3 & 74.6 & 83.8 & 88.8 & 2.47\% & 82.8 & 85.3 & 86.8 & 84.2 & 87.5 & 89.5 & 1.69\% \\
      desk\_changing & 71.3 & 77.2 & 80.4 & 69.7 & 76.8 & 80.4 & 73.5 & 81.6 & 85.8 & 74.1 & 82.5 & 87.0 & 0.82\% & 61.2 & 62.9 & 63.7 & 76.2 & 82.3 & 85.5 & 24.51\% \\
      einstein & 57.1 & 62.0 & 64.8 & 66.7 & 72.4 & 75.8 & 67.8 & 74.1 & 77.8 & 74.7 & 82.2 & 86.5 & 10.18\% & 36.6 & 38.2 & 39.4 & 74.0 & 81.5 & 85.9 & 102.19\% \\
      einstein\_GLC & 42.3 & 46.1 & 48.4 & 51.7 & 57.4 & 60.7 & 51.3 & 56.7 & 60.3 & 51.6 & 57.1 & 60.7 & 0.58\% & 35.5 & 41.0 & 45.9 & 49.2 & 56.1 & 62.5 & 38.59\% \\
      mannequin & 76.2 & 80.6 & 83.0 & 80.1 & 84.9 & 87.5 & 80.2 & 85.3 & 88.0 & 84.2 & 90.3 & 93.6 & 4.99\% & 59.5 & 60.9 & 61.6 & 74.2 & 77.9 & 79.8 & 24.71\% \\
      mannequin\_face & 69.1 & 71.0 & 71.9 & 73.2 & 76.3 & 77.9 & 73.4 & 76.8 & 78.5 & 76.4 & 79.9 & 81.7 & 4.09\% & 53.9 & 54.3 & 54.5 & 76.7 & 80.3 & 82.2 & 42.30\% \\
      planar & 67.7 & 78.9 & 84.5 & 65.6 & 79.5 & 86.7 & 67.1 & 81.6 & 88.9 & 68.6 & 82.9 & 90.3 & 2.24\% & 62.9 & 70.6 & 74.4 & 71.8 & 80.7 & 85.2 & 14.15\% \\
      plant & 78.5 & 82.2 & 84.1 & 74.4 & 80.2 & 83.0 & 71.9 & 78.5 & 81.7 & 80.2 & 86.3 & 89.4 & 11.54\% & 86.5 & 88.9 & 90.0 & 89.4 & 91.6 & 92.7 & 3.35\% \\
      plant\_scene & 72.6 & 77.2 & 79.6 & 71.0 & 76.4 & 79.2 & 71.8 & 77.7 & 80.7 & 79.3 & 85.7 & 88.9 & 10.45\% & 44.2 & 45.2 & 45.7 & 79.0 & 86.6 & 90.6 & 78.73\% \\
      sfm\_lab\_room & 87.7 & 93.8 & 96.9 & 90.6 & 95.3 & 97.6 & 86.4 & 93.3 & 96.6 & 88.8 & 94.4 & 97.2 & 2.78\% & 92.0 & 94.4 & 95.6 & 96.0 & 98.0 & 99.0 & 4.35\% \\
      sofa & 69.2 & 76.9 & 81.4 & 70.8 & 79.6 & 84.2 & 70.8 & 79.6 & 84.1 & 74.2 & 83.6 & 88.7 & 4.80\% & 52.9 & 54.2 & 54.8 & 74.3 & 81.8 & 85.6 & 40.45\% \\
      table & 65.3 & 68.8 & 70.6 & 65.3 & 69.4 & 71.4 & 70.5 & 75.3 & 77.7 & 73.5 & 78.8 & 81.5 & 4.26\% & 28.6 & 29.4 & 29.8 & 71.3 & 76.5 & 79.1 & 149.30\% \\
      vicon\_light & 69.4 & 76.0 & 79.6 & 72.7 & 81.1 & 85.6 & 73.2 & 82.0 & 86.6 & 74.2 & 83.2 & 87.8 & 1.37\% & 66.5 & 70.1 & 72.1 & 77.1 & 84.6 & 88.6 & 15.94\% \\
      large\_loop & 69.2 & 75.1 & 78.1 & 69.9 & 75.8 & 78.8 & 73.2 & 79.3 & 82.4 & 78.0 & 85.8 & 90.0 & 6.56\% & 42.5 & 45.1 & 46.4 & 77.4 & 85.9 & 90.5 & 82.12\% \\
      \midrule
      Overall & 69.7 & 75.8 & 79.0 & 70.9 & 77.8 & 81.4 & 71.5 & 79.0 & 83.0 & 75.1 & 82.8 & 86.9 & 5.03\% & 58.2 & 61.2 & 62.8 & 76.4 & 82.3 & 85.6 & 31.27\% \\
      \bottomrule
    \end{tabular}
  }
  \label{tab:eth3d}
\end{table*}

\subsubsection{Results}
In our experiments, we did not alter the theoretical framework or design principles of the original iMatching method. Consequently, the performance metrics of our implementation are expected to closely match those reported in \cite{zhan2024imatching}. Our experimental results confirm this expectation, underscoring the compatibility and effectiveness of our integration. We report the accuracy on the KITTI360 dataset in \tref{tab:kitti}, with the best results highlighted in bold. Specifically, our implementation of iDKM achieves identical accuracy to that reported in \cite{zhan2024imatching}. We also show the zero-shot visualization results on the Waymo Open Dataset \cite{waymo} in \fref{fig:waymo_dkm}. Notably, our implementation of iASpan outperforms the previously reported results by 3.2\%, reflecting a significant improvement of 17\% over the pretrained ASpanFormer baseline. The evaluation results on the ETH3D-SLAM dataset are presented in \tref{tab:eth3d}. Although individual scenes exhibit some variation due to the inherent variance of smaller samples, the overall accuracy aligns closely with \cite{zhan2024imatching}. The entries labeled “iCAPS” and “iDKM” indicate accuracies of CAPS and DKM after training through self-supervised learning, yielding notable improvements of 5.03\% and 27.3\%, respectively. Our results show that our proposed framework could be used in complex learning framework and can adapt to SOTA models.

\subsection{Generalization to Pose Graph Optimization}
\label{sec:pgo}

\begin{table}[t]
    \centering
    \caption{Performance comparison for Pose Graph Optimization.}
    \label{table:pgo}
  \setlength{\tabcolsep}{0.26em}
    \resizebox{\linewidth}{!}{
    \begin{tabular}{lcccc}
        \toprule
        & \multicolumn{2}{c}{\texttt{parking-garage}} & \multicolumn{2}{c}{\texttt{sphere-a}} \\
        Method & Error$\downarrow$ & Time (s) $\downarrow$ & Error$\downarrow$ & Time (s) $\downarrow$ \\
        \midrule
        PyPose \cite{wang2023pypose} & 6.2793e-01 & 9m29s & 6.3789e+04 & 18m28s \\
        Ceres Solver \cite{ceres} & 6.34188e-01 & 0.81 &  6.3786e+04 & 14.90 \\
        Ours (Cholesky) & 6.34347e-01 & 0.86 & 6.3789e+04 & 1.69  \\
        Ours (PCG) & 6.34435e-01 & 33.82 & 6.3789e+04 & 17.66   \\
        \bottomrule
    \end{tabular}
    }
\end{table}

\begin{figure}[t]
    \centering
    \includegraphics[width=1.0\linewidth]{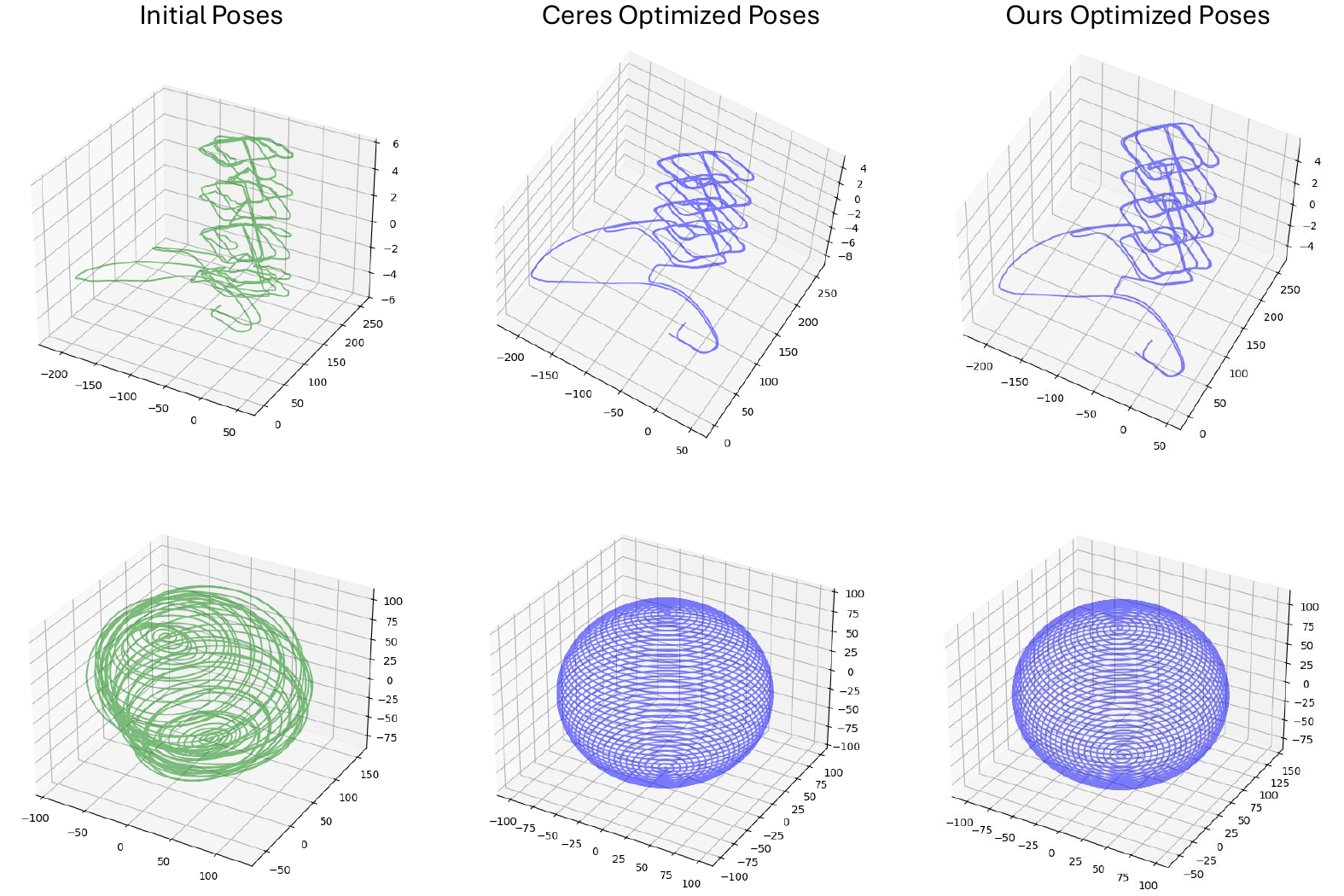}
     \caption{\textbf{Qualitative Comparison of Pose Optimization Results.} Top row: Trajectory visualization of camera poses from the parking garage dataset. Bottom row: Pose graph optimization on the synthetic "sphere" dataset. Each column shows (left) the initial poses before optimization, (middle) results from Ceres Solver, and (right) results from our proposed eager-mode BA framework. Both optimization methods successfully refine the poses.}
     \label{fig:pgo}
     \vspace{-2em}
\end{figure}

To validate the generality of our framework beyond BA, we evaluate its performance on PGO, as formulated in \sref{sec:pgo_method}, on the benchmark problems: \texttt{sphere-a} and \texttt{parking-garage}; \texttt{sphere-a} is a synthetic sample released in \cite{g2o}, and \texttt{parking-garage} is a real-world sample \cite{OpenSLAMVertigo}. We compare our framework against PyPose \cite{wang2023pypose} and Ceres Solver \cite{ceres}, all used for PGO. Results are summarized in Table \ref{table:pgo}, and \fref{fig:pgo} shows qualitative results. PyPose represents a baseline LM optimizer without sparsity support. In solving PGO, our method is \textbf{659$\times$} faster than PyPose, compared to its latest version at the submission of this work. Additionally, our method achieves comparable accuracy with Ceres Solver. \texttt{parking-garage} is a small sample with only 1661 nodes and 6275 edges, so our efficiency is bounded by the Python interpreter and shows a comparable runtime to Ceres Solver. The larger \texttt{sphere-a} sample has 2500 nodes and 9799 edges with larger noise. Our method demonstrates a speedup of 8.87$\times$. This experiment demonstrates that our proposed framework is adaptable to more types of problems with different compute graphs.

\section{Conclusions \& Discussions}

We introduced a highly extensible, efficient, and scalable BA library in the eager execution mode, fully compatible with PyTorch. This library leverages GPU acceleration, a novel sparsity-aware AutoDiff strategy, and specialized sparse linear algebra operations, substantially outperforming traditional BA frameworks such as Ceres Solver, GTSAM, and \gTwoO. Our comprehensive evaluation across benchmark datasets and deep-learning-driven SfM pipelines demonstrated remarkable improvements, achieving up to hundreds of times speed-up on large-scale problems while maintaining high accuracy.

Furthermore, we showcased the seamless integration of our library into modern deep learning workflows, notably enhancing training efficiency and simplicity in complex pipelines like iMatching and VGGT. Its generalization to PGO also illustrated the versatility and adaptability of our approach beyond BA. By supporting PyTorch's intuitive eager mode and maintaining native tensor-based interfaces, our library significantly lowers the entry barrier for researchers to rapidly prototype, experiment, and debug.

Despite these advantages, several areas remain open for potential improvement.
First, our current implementation favors GPU execution. CPU-oriented optimization techniques such as multi-threading with Intel OneTBB and SIMD instructions could be introduced.
Additionally, Python's automatic garbage collection and PyTorch's dynamic tensor management, while convenient, lead to higher memory overhead than statically compiled C++-based libraries. Manual buffer preallocation, TorchDynamo, or TorchScript compilation could bring additional runtime and memory savings.
Lastly, expanding compatibility to frameworks like JAX or MLX, extending support to diverse optimization tasks such as robotic manipulation or motion planning, and adding loss functions and reusable optimization templates supporting those tasks (e.g., dogleg, L-BFGS) could enhance broader accessibility and applicability.

{
\small
\bibliographystyle{IEEEtran}
\bibliography{pp06, publications, imat-main, main, vggt}
}

\appendices

\section{Background of iMatching}
\label{x:imatching}
\subsubsection{Background}
iMatching finetunes pretrained feature matching models using unannotated videos without relying on any form of ground-truth supervision. It achieves this by utilizing the BA reprojection error $\mathbf{r}$ defined in Eq. \eqref{eq:nls} as a self-supervised loss to train the feature matching model, while simultaneously refining the 3D scene geometry through multiview observations. This strategy promotes a mutual correction between the feature matching network and the BA module, contrasting with traditional BA methods that provide no direct feedback to the feature extraction process.
To enable this self-supervised training, iMatching formulates the problem as a bilevel optimization. At the lower level, the BA module optimizes camera poses $\bm{\zeta}$ and 3D landmark positions $\mathbf{p}$ to minimize the reprojection error $\mathbf{r}$. At the upper level, the parameters of the feature matching network are updated to reduce $\mathbf{r}$, leveraging the optimized camera poses and landmarks. This integrated design allows BA to supply the geometric supervision required for the feature matching network to learn accurate correspondences without using ground-truth.

\subsubsection{Training Method}
In each training iteration, the pretrained model predicts feature correspondences for a short video clip of 4-6 frames. The features are first used for estimating camera poses and 3D landmarks. The BA module then refines the estimations by minimizing the reprojection error using the LM optimizer. The final reprojection error is then used for updating the model weights in the upper level.

\section{CUDA Graphs for PCG Acceleration}
\label{x:cudagraph}
For readability, our iterative PCG is implemented entirely in PyTorch's Python syntax. However, it involves repeated launches of the same CUDA kernels for SpMV, vector updates, and scalar reductions in all iterations. This traditionally incurs overhead due to the Python interpreter and PyTorch kernel launches. To eliminate this overhead, we incorporate CUDA graph capture and replay \cite{nvidia_cuda_graphs_2025}. During the initial PCG iteration, we record a CUDA graph capturing the entire PCG iteration, including all required kernel launches and memory operations. In subsequent iterations, we replay this captured CUDA graph. This strategy reduces kernel launch overhead to further approach the theoretical peak GPU utilization.



\begin{IEEEbiography}
[{\includegraphics[width=1in,height=1.25in,clip,keepaspectratio]{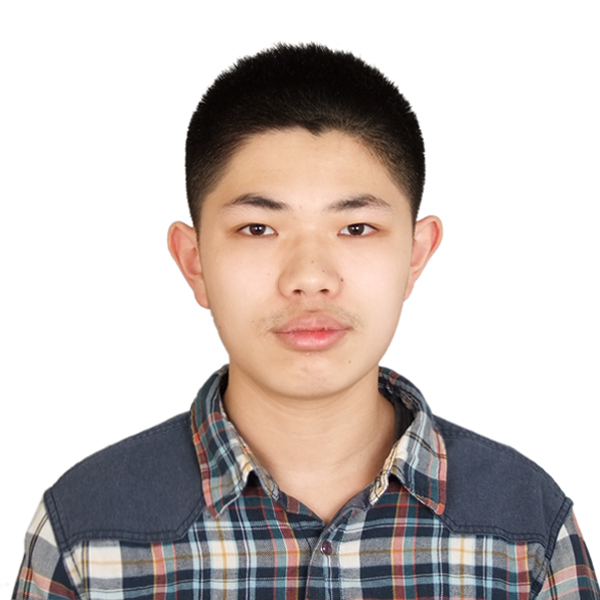}}]
{Zitong Zhan} is a Ph.D. student with the Department of Computer Science and Engineering at the University at Buffalo, State University of New York. He received his B.S. in Computer Science and Mathematics from the University of Wisconsin-Madison in 2021 and his M.S. in Computer Science from the University of Illinois at Urbana-Champaign in 2023. His research interests include building system for computer vision and robotics, such as structure-from-motion and simultaneous localization and mapping. 
\end{IEEEbiography}

\begin{IEEEbiography}[{\includegraphics[width=1in,height=1.25in,clip,keepaspectratio]{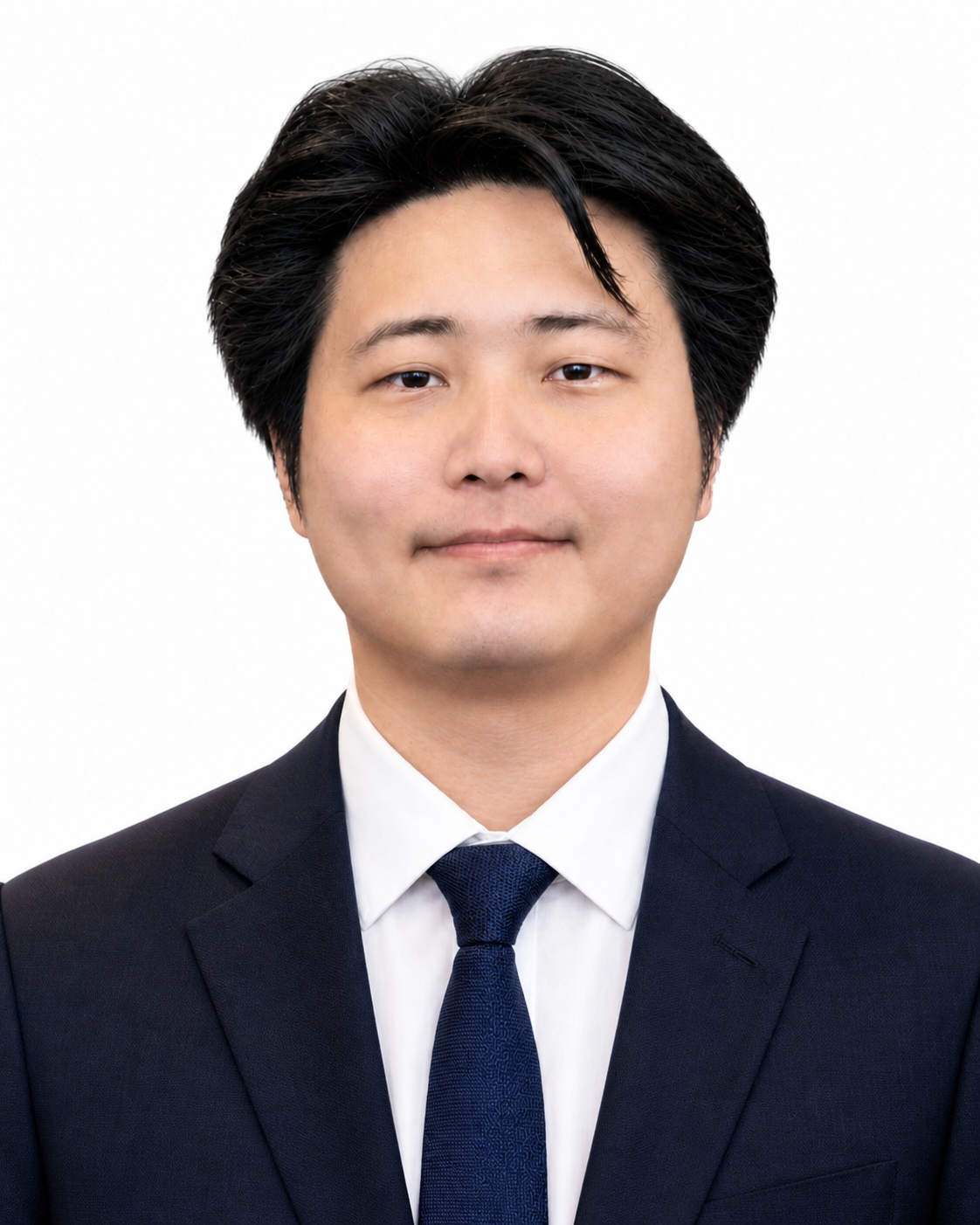}}]
{Huan Xu} received the M.S. degree in Computer Science from Georgia Institute of Technology in 2025 and received his B.S. in Computer Science, Mathematics, and Statistics from the University of Wisconsin-Madison in 2023. His research interests include accelerated computing, GPU performance optimization, and software systems for massively parallel architectures.
\end{IEEEbiography}

\begin{IEEEbiography}
[{\includegraphics[width=1in,height=1.25in,clip,keepaspectratio]{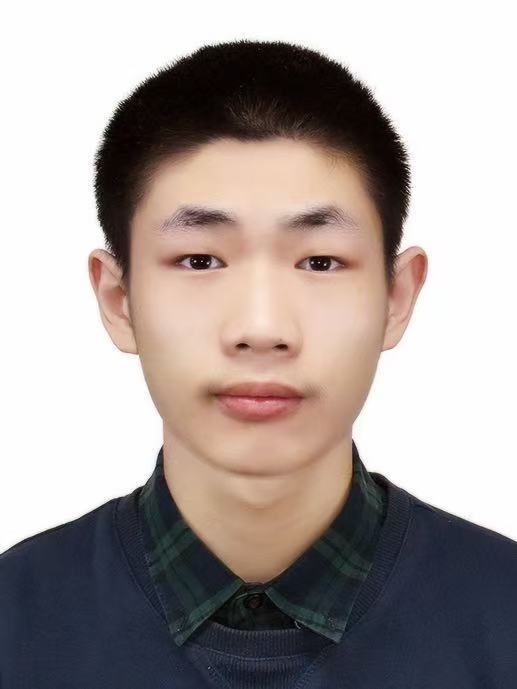}}]
{Zihang Fang} is an undergraduate student at Purdue University. His research interests include optimization and control for robotic systems.
\end{IEEEbiography}

\begin{IEEEbiography}[{\includegraphics[width=1in,height=1.25in,clip,keepaspectratio]{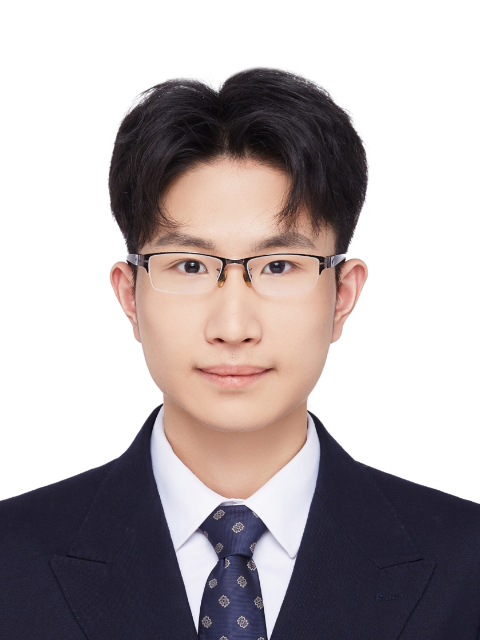}}]
{Xinpeng Wei} received the M.S. degree in Computer Science from Georgia Institute of Technology in 2025 and B.Eng. degree in Software Engineering from Shanghai Jiao Tong University in 2023. His research interests include building systems for large-scale machine learning like pre-training and reinforcement learning of large language models.
\end{IEEEbiography}

\begin{IEEEbiography}
[{\includegraphics[width=1in,height=1.25in,clip,keepaspectratio]{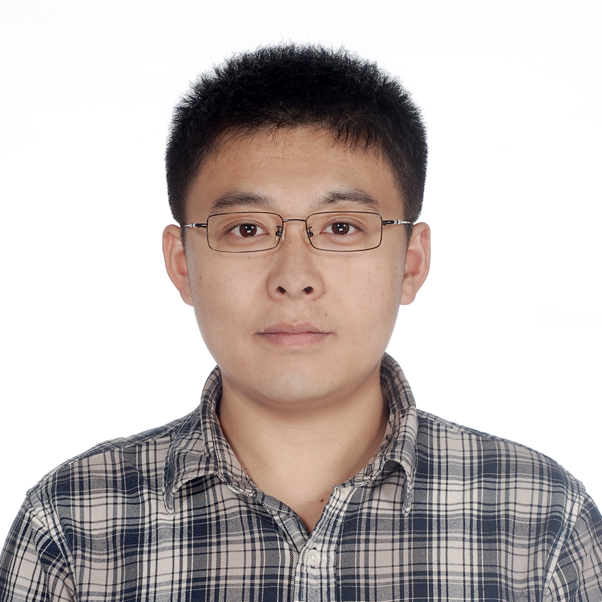}}]
{Yaoyu Hu} received the Ph.D. degree from Shanghai Jiao Tong University, Shanghai, China, in 2017. He is currently a Senior Project Scientist with the Robotics Institute (RI), Carnegie Mellon University (CMU), Pittsburgh, PA, USA, where he is involved in learning-based visual odometry and multimodal sensor fusion. His research interests include general scene understanding, such as class agnostic object detection, tracking, and 3-D scene reconstruction.
\end{IEEEbiography}

\begin{IEEEbiography}[{\includegraphics[width=1in,height=1.25in,clip,keepaspectratio]{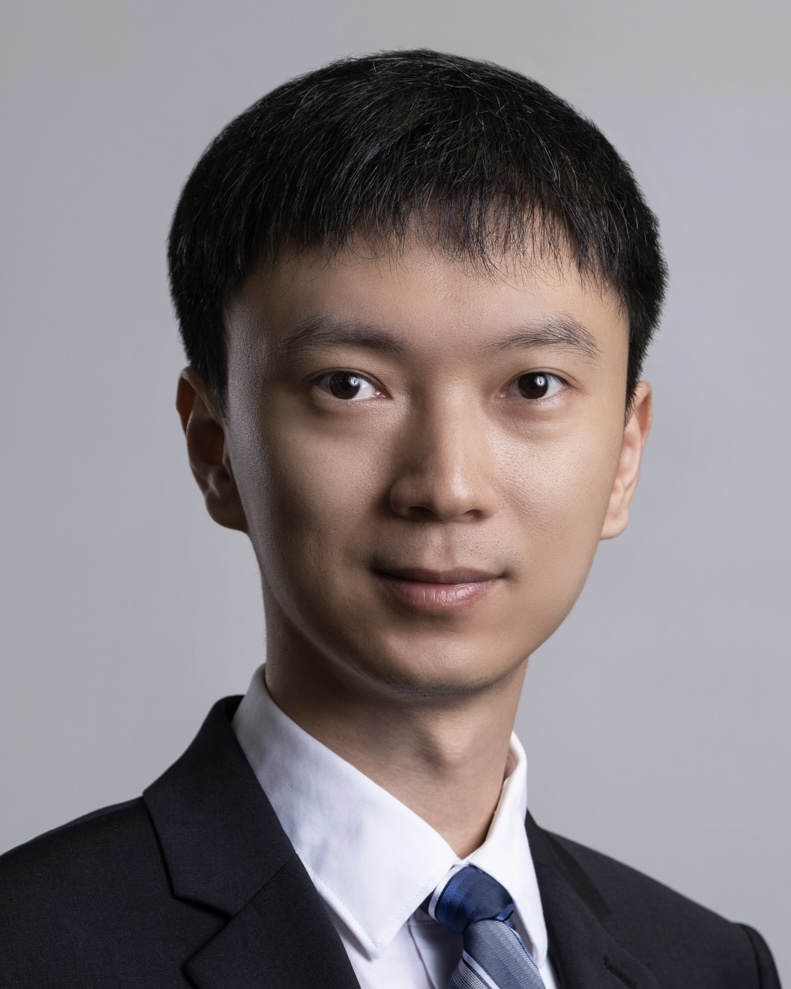}}]
{Chen Wang} (Senior Member, IEEE) is an Assistant Professor in the Department of Computer Science and Engineering at the University at Buffalo (UB), where he leads the Spatial AI and Robotics (SAIR) Lab. He received the B.Eng. degree in Electrical Engineering from Beijing Institute of Technology (BIT) in 2014 and the Ph.D. degree in Electrical Engineering from Nanyang Technological University (NTU), Singapore, in 2019. From 2019 to 2022, he was a Postdoctoral Fellow at the Robotics Institute, Carnegie Mellon University (CMU).

Dr. Wang's research focuses on Spatial AI and robotics. He has served as an Associate Editor for IEEE Transactions on Robotics (T-RO), International Journal of Robotics Research (IJRR), and IEEE Robotics and Automation Letters (RA-L), and as an Area Chair for the IEEE/CVF Conference on Computer Vision and Pattern Recognition (CVPR), IEEE International Conference on Robotics and Automation (ICRA), and the Conference on Neural Information Processing Systems (NeurIPS). He also served as an Associate Co-chair of the IEEE RAS Technical Committee on Computer and Robot Vision.
\end{IEEEbiography}

\vfill

\end{document}